\documentclass[11pt,a4paper]{article}

\usepackage[UKenglish]{babel}
\usepackage[T1]{fontenc}
\usepackage[utf8]{inputenc} 
\usepackage{lmodern}
\usepackage{microtype}
\usepackage[a4paper,margin=1in]{geometry}

\usepackage{mathtools}
\usepackage{amssymb}
\usepackage{amsthm}
\usepackage{mathrsfs}
\usepackage{cmll}
\usepackage{comment}

\usepackage{graphicx}
\usepackage{subcaption}
\usepackage{booktabs}

\usepackage{tikz}
\usepackage{tikzit}     
\usepackage{circuitikz}
\usetikzlibrary{circuits.ee.IEC}
\input{styles.tikzdefs}   

\tikzstyle{discarding}=[fill=white, draw=black, shape=circle, style=upground]
\tikzstyle{wide_discarding}=[fill=white, draw=black, shape=circle, style=upground, yscale=1.3]
\tikzstyle{smalldiscarding}=[fill=white, draw=black, style=upground, scale=0.75]
\tikzstyle{backdiscard}=[fill=white, draw=black, shape=circle, style=downground, scale=0.5]
\tikzstyle{smallbackdiscard}=[fill=white, draw=black, shape=circle, style=downground, scale=0.75]
\tikzstyle{state}=[fill=white, draw=black, style=triang, tikzit shape=rectangle]
\tikzstyle{effect}=[fill=white, draw=black, style=triangdag]
\tikzstyle{black_dot}=[style=dot, fill=black]
\tikzstyle{white_dot}=[style=dot, fill=white]
\tikzstyle{qblack_dot}=[style=ddot, fill=black]
\tikzstyle{qwhite_dot}=[style=ddot, fill=white]
\tikzstyle{whitephase}=[style=wphase dot, fill=white]

\tikzstyle{dottededge}=[-, dash pattern=on 2pt off 1pt]
\tikzstyle{thin_edge}=[-, style=thin]
\tikzstyle{thick_edge}=[-, style=thick]
\tikzstyle{thicker_edge}=[-, style=thicker]
\tikzstyle{double edge}=[-, style=doubled, draw=black, tikzit draw={rgb,255: red,18; green,168; blue,191}]
\tikzstyle{arrow}=[->]
\tikzstyle{new edge style 1}=[-, draw={rgb,255: red,242; green,233; blue,206}, fill={rgb,255: red,242; green,233; blue,206}]
\tikzstyle{morphism_shade}=[-, draw=black, fill={rgb,255: red,242; green,233; blue,206}, line join=bevel]
\tikzstyle{supermap_shade}=[-, fill={rgb,255: red,216; green,215; blue,242}, draw=black, line join=bevel]
\tikzstyle{hole_shade}=[-, fill=white, draw=black, line join=bevel]
\tikzstyle{new edge style 2}=[-, draw={rgb,255: red,14; green,188; blue,83}]
\tikzstyle{green}=[-, fill=none, draw={rgb,255: red,0; green,106; blue,106}]
\tikzstyle{green_dotted}=[-, draw={rgb,255: red,0; green,106; blue,106}, dash pattern=on 1pt off 0.7pt]

\usepackage[hidelinks]{hyperref}
\usepackage[nameinlink,noabbrev]{cleveref}

\usepackage{thmtools}
\usepackage{thm-restate}

\providecommand{\ccsdesc}[2][]{}

\providecommand{\authorrunning}[1]{}
\providecommand{\titlerunning}[1]{}
\providecommand{\Copyright}[1]{}

\theoremstyle{plain}
\newtheorem{theorem}{Theorem}[section]
\newtheorem{lemma}[theorem]{Lemma}
\newtheorem{proposition}[theorem]{Proposition}
\newtheorem{corollary}[theorem]{Corollary}

\theoremstyle{definition}
\newtheorem{definition}[theorem]{Definition}

\theoremstyle{remark}

\newcommand{\morph}[1]{\xrightarrow{#1}}

\newcommand{\hypdec}{\mathtt{hypdec}}
\newcommand{\dec}{\mathtt{dec}}

\newcommand{\realsgeq}{\mathbb{R}_{\ge 0}}

\title{Agent policies from higher-order causal functions}
\author{%
  Matt Wilson\\
  {\small Universit\'e Paris-Saclay, CNRS, ENS Paris-Saclay, Inria, CentraleSup\'elec, Laboratoire M\'ethodes Formelles}\\
  {\small\texttt{matthew.wilson@centralesupelec.fr}}
}
\date{}       

\begin{document}
\maketitle

\begin{abstract}
We establish a correspondence between equivalence classes of agent-state policies for deterministic POMDPs and one-input process functions (the classical-deterministic limit of higher-order quantum operations). We use this correspondence to build a bridge between the agent-environment interaction in artificial intelligence, causal structure in the foundations of physics, and logic in computer science.
We construct a *-autonomous category $\mathbf{PF}$ of types which supports an interpretation of one-step evaluation of policies, and multi-agent observation constraints, into cuts and monoidal products. 
In terms of types, we develop the correspondence further by identifying observation-independent decentralised POMDPs as the natural domain for the multi-input process functions used in the foundations of physics to model indefinite causality.
We then prove a strict separation between general multi-input process function and definite-ordered process function performance on such $\dec$-POMDPs, by finding an instance for which policies utilizing an indefinite causal structure can achieve greater finite-horizon rewards than policies which are restricted to a fixed background causal structure.
\end{abstract}

\section{Introduction}
Agency, the capability of an entity to act upon and receive information from its surroundings, is a fundamental notion in both artificial intelligence and the foundations of physics.  In AI, agents interact with partially observable environments to maximise cumulative reward, forming the basis for planning and learning for single and multi-agent systems \cite{lu2023reinforcementlearningbitbit, OliehoekAmato2016decPOMDP}. In the informational foundations of physics, “agents in laboratories” are modelled as local operations inserted into a spacetime environment, formalised by higher-order quantum  \cite{taranto2025higherorderquantumoperations, barrett2020quantumcausalmodels, oreshkov, chiribella_switch} and classical \cite{Baumeler_2016} processes. Whilst the agent-environment interaction is central to both fields, their formalisations of the concept have developed independently and no direct mathematical correspondence between them has been established.

A bridge between these two models for the agent-environment interaction has the possibility to bring new ideas to both fields. First, regarding causality and spacetime, higher-order processes allow causal (even indefinite-causal) structure to be treated as a resource for communication and computation \cite{Baumeler_2016, Ebler_2018, Gu_rin_2016}.  A mapping to planning and learning agents opens the door to consideration of multi-agent tasks in which optimal causal and indefinite causal strategies might exist and further be discovered and learned. Second, regarding the quantization of agents, interpreting agents as higher-order processes suggests a principled route to quantum generalizations of planning and learning agents \cite{acampora2025quantumcomputingartificialintelligence, PhysRevA.90.032311, saldi2024quantummarkovdecisionprocesses} in terms of higher-order quantum operations, giving a stable formal framework for the analysis of quantum advantage in multi-agent systems.   

From the formal and logical perspective, the compositional and logical tools developed for higher-order quantum maps \cite{Bisio_2019, Apadula_2024, kissinger_caus,  simmons2024completelogiccausalconsistency, simmons2022higherordercausaltheoriesmodels, hoffreumon2024projectivecharacterizationhigherorderquantum, jencova2024structurehigherorderquantum} and their generalisation to arbitrary monoidal categories \cite{hefford_supermaps,hefford2025bvcategoryspacetimeinterventions, wilson2025quantumsupermapscharacterizedlocality} might be utilised in the context of such an identification to provide tools for reasoning about composite multi-agent systems. Conversely, already existing compositional and logical tools for modelling aspects of reward-seeking agents and environments in categorical cybernetics \cite{Capucci_2022, Hedges_2023, Hedges_2025} and open game theory \cite{ghani2018compositionalgametheory} could be lifted to the quantum domain to bring new techniques and ways of thinking about quantum reinforcement learning \cite{chen2024introductionquantumreinforcementlearning} and quantum game theory \cite{Gutoski_2007}.

In this paper we give a precise correspondence between deterministic agent-state policies \cite{lu2023reinforcementlearningbitbit, dong2021simpleagentcomplexenvironment, sinha2024agentstatebasedpoliciespomdps, sinha2024periodicagentstatebasedqlearning} for deterministic POMDPs and process functions \cite{Baumeler_2016, Baumeler_2021, dourdent2025paradoxfreeclassicalnoncausalityunambiguous}, the classical deterministic limit of higher-order quantum operations, and extend the identification to the case of decentralised multi-agent systems. In this identification, the evaluation of a policy on a POMDP is transformed into the evaluation of a higher-order function representing the policy on a lower-order function representing the POMDP.
Using the defining fixed-point criterion for process functions, we generalize them to a broad class of \textit{types}, and show that they can be arranged into a $*$-autonomous category \cite{Barr1979Autonomous}, for which the associated tensors allow for a purely type-theoretic expression of decentralization and observation-independence \cite{OliehoekAmato2016decPOMDP}, as well as indefinite causal structure \cite{oreshkov, chiribella_switch}. 
Noting that observation-independent decentralised POMDPs coincide with the natural domain for multi-input process functions, we discover a proof-of-principle POMDP based on causal games \cite{Baumeler_2016} in which policies utilising decentralised indefinite causal structures achieve strictly higher finite-horizon and asymptotic rewards than decentralised definite-ordered counterparts.

\section{Agent-state policies}
We will work with deterministic partially observable Markov decision processes (POMDPs) \cite{bonet2012deterministicpomdpsrevisited}.  A deterministic POMDP is defined by the tuple $\langle S,A,\Omega,\mathcal{T},\mathcal{O},\mathcal{R} \rangle$ where $S$ is a set of states, $A$ a set of available actions, $\Omega$ a set of observations and
$\mathcal{T}: A\times S \rightarrow  S$, $\mathcal{O}: A\times S \rightarrow  \Omega$ and $\mathcal{R}: A\times S \rightarrow  \realsgeq$ are deterministic functions specifying the state update, the observation, and the reward associated to performing an action in a given state.  From here on we will drop $S,A,\Omega$ and simply write $\langle \mathcal{T},\mathcal{O},\mathcal{R}\rangle$ whenever the meaning is clear.

It is imagined that an agent interacting with a POMDP does not have access to the state of the environment, and so to act optimally must retain a memory of the history of its past interactions.  
Accordingly, a policy for a POMDP is typically taken to mean the specification of a function
$ \pi : \texttt{Hist}(A \times \Omega) \rightarrow A $, where $ \texttt{Hist}(X)$ is the set of lists of elements of $X$ interpreted as a history of action-observation pairs. It is imagined, that each time an agent takes an action $a$ and makes an observation $o$ then for the next round the action-observation pair $(a,o)$ is recorded by appending it to the current history $H$. 

In this article we will work with agent-states, which generalize action-observation histories to allow for any agent to retain an abstract memory, representing a suitable compression of the history of past observations \cite{lu2023reinforcementlearningbitbit, dong2021simpleagentcomplexenvironment, sinha2024agentstatebasedpoliciespomdps, sinha2024periodicagentstatebasedqlearning}. 
Whilst in these works, agent policies and updates are permitted to be stochastic, we will work entirely within the deterministic picture for simplicity (in particular, to avoid the problems of infinite dimensionality which would arise when maintaining an arbitrary-length history).

Accordingly, we say that a deterministic agent-state policy is specified by two functions:
\begin{enumerate}
  \item A \emph{policy} $\pi: M\rightarrow  A$ that selects an action based on the current memory state.
  \item A \emph{memory update} $\mathcal{U}: M\times A\times \Omega \rightarrow  M$ that updates the agent’s memory based on the previous memory, the chosen action, and the observed outcome.
\end{enumerate}
We will drop the word deterministic whenever the context is clear. Note that an agent-state policy includes not just the policy function but also the updating mechanism for memory (which is left implicit as list concatenation in purely history-based policies).  

Let us now move to the multi-agent setting, in particular the decentralised setting \cite{bernstein2013complexitydecentralizedcontrolmarkov, OliehoekAmato2016decPOMDP} in which multiple agents act locally, using their own memories to choose actions and their own observations to update their memories; 

\begin{definition}[$n$-party deterministic $\dec$-POMDP]
\label{definition:n_dec_pomdp}
We consider a global environment state space $S$ and families $A_i, \Omega_i$ of local action and observation spaces respectively, a \emph{deterministic $n$-party $\dec$-POMDP} is, simply a POMDP with state set $S$, action set $\prod_{i=1}^n A_i$, and observation set $ \prod_{i=1}^n \Omega_i$.
\end{definition}

\newcommand{\caplinepad}{\vphantom{\rule{0pt}{\baselineskip}}} 

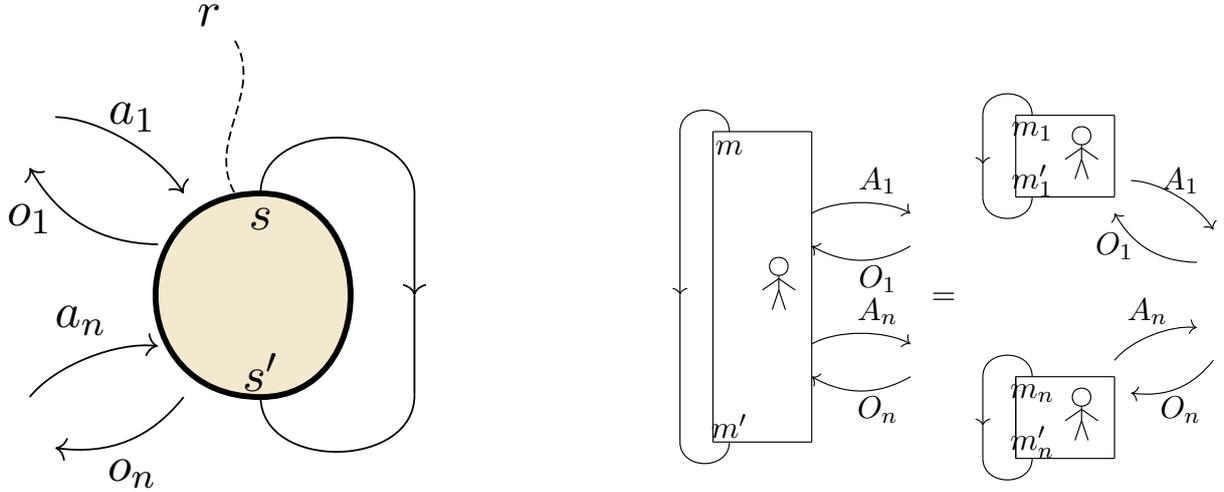
\begin{figure}[h]
  \centering
  \begin{subfigure}[b]{0.36\linewidth}
    \centering
    \resizebox{\linewidth}{!}{\begin{tikzpicture}
	\begin{pgfonlayer}{nodelayer}
		\node [style=none] (87) at (5.25, 0.5) {};
		\node [style=none] (88) at (4, 1.25) {};
		\node [style=none] (89) at (5, 0) {};
		\node [style=none] (90) at (3.75, 0.75) {};
		\node [style=none] (91) at (4.75, 1.25) {$a_1$};
		\node [style=none] (92) at (3.75, 0.25) {$o_1$};
		\node [style=none] (121) at (6, 0.5) {};
		\node [style=none] (122) at (6, -1.5) {};
		\node [style=none] (123) at (6, 0.5) {};
		\node [style=none] (124) at (6, -1.5) {};
		\node [style=none] (127) at (6, 0.5) {};
		\node [style=none] (128) at (6, -1.5) {};
		\node [style=none] (140) at (6, 0.5) {};
		\node [style=none] (141) at (6, -1.5) {};
		\node [style=none] (142) at (6, 0.5) {};
		\node [style=none] (143) at (6, 0.5) {};
		\node [style=none] (144) at (6, -1.5) {};
		\node [style=none] (145) at (6, -1.5) {};
		\node [style=none] (146) at (5.75, 0.5) {};
		\node [style=none] (149) at (6, 0.25) {$s$};
		\node [style=none] (150) at (6, -1.25) {$s'$};
		\node [style=none] (188) at (5, -1) {};
		\node [style=none] (189) at (3.75, -1.5) {};
		\node [style=none] (190) at (5.25, -1.5) {};
		\node [style=none] (191) at (4, -2) {};
		\node [style=none] (192) at (4.25, -0.75) {$a_n$};
		\node [style=none] (193) at (4.75, -2.25) {$o_n$};
		\node [style=none] (195) at (5.75, 0.5) {};
		\node [style=none] (196) at (5.75, 2) {};
		\node [style=none] (197) at (5.5, 2.25) {$r$};
		\node [style=none] (202) at (7.5, 0.5) {};
		\node [style=none] (203) at (7.5, -1.475) {};
		\node [style=none] (204) at (6, 0.5) {};
		\node [style=none] (205) at (6, -1.5) {};
		\node [style=none] (206) at (6, 0.5) {};
		\node [style=none] (207) at (6, -1.5) {};
		\node [style=none] (209) at (6, 0.5) {};
		\node [style=none] (210) at (6, -1.5) {};
		\node [style=none] (215) at (7.5, -0.5) {};
	\end{pgfonlayer}
	\begin{pgfonlayer}{edgelayer}
		\draw [style=arrow, bend left, looseness=0.75] (88.center) to (87.center);
		\draw [style=arrow, bend left] (89.center) to (90.center);
		\draw [style={morphism_shade}] (121.center)
			 to (123.center)
			 to [bend right=90, looseness=1.75] (124.center)
			 to (122.center)
			 to [bend right=90, looseness=1.50] cycle;
		\draw [style=dottededge, bend right=90, looseness=1.75] (140.center) to (141.center);
		\draw [style=double edge] (142.center) to (143.center);
		\draw [style=double edge, bend left=270, looseness=1.75] (143.center) to (144.center);
		\draw [style=double edge] (144.center) to (145.center);
		\draw [style=double edge, bend right=90, looseness=1.50] (145.center) to (142.center);
		\draw [style=arrow, bend left, looseness=0.75] (189.center) to (188.center);
		\draw [style=arrow, bend left] (190.center) to (191.center);
		\draw [style=dottededge, in=120, out=-60] (196.center) to (195.center);
		\draw [bend left=90, looseness=1.25] (204.center) to (202.center);
		\draw [bend right=90, looseness=1.25] (205.center) to (203.center);
		\draw [style=arrow] (202.center) to (215.center);
		\draw (215.center) to (203.center);
	\end{pgfonlayer}
\end{tikzpicture}}
    \caption{A decentralised POMDP ($\dec$-POMDP) accepts actions from multiple independent parties and returns them each a separate observation.}
    \label{fig:lics_dec_pomdp_cartoon}\label{fig:lics_dec_pomdp_cartoon_1}
  \end{subfigure}\hfill
  \begin{subfigure}[b]{0.45\linewidth}
    \centering
    \resizebox{\linewidth}{!}{$\begin{tikzpicture}
	\begin{pgfonlayer}{nodelayer}
		\node [style=none] (68) at (-0.75, 2.5) {};
		\node [style=none] (69) at (0.75, 2.5) {};
		\node [style=none] (70) at (0.75, -2.25) {};
		\node [style=none] (71) at (-0.75, -2.25) {};
		\node [style=none] (74) at (0.5, -0.25) {};
		\node [style=none] (75) at (0.5, 0.5) {};
		\node [style=none] (76) at (0.5, -0.25) {};
		\node [style=none] (77) at (0.5, 0.5) {};
		\node [style=none] (78) at (0.25, 0.3) {};
		\node [style=none] (79) at (0.25, 0.075) {};
		\node [style=none] (80) at (0.35, -0.225) {};
		\node [style=none] (81) at (0.15, -0.225) {};
		\node [style=none] (82) at (0.5, 0.075) {};
		\node [style=none] (83) at (0, 0.075) {};
		\node [style=none] (84) at (0.25, 0.25) {};
		\node [style=none] (85) at (0.25, 0.575) {};
		\node [style=none] (86) at (0, 0.075) {};
		\node [style=none] (87) at (2.25, 1.25) {};
		\node [style=none] (88) at (0.75, 1.25) {};
		\node [style=none] (89) at (2.25, 0.75) {};
		\node [style=none] (90) at (0.75, 0.75) {};
		\node [style=none] (91) at (1.75, 1.75) {$A_1$};
		\node [style=none] (92) at (1.75, 0.25) {$O_1$};
		\node [style=none] (188) at (2.25, -0.75) {};
		\node [style=none] (189) at (0.75, -0.75) {};
		\node [style=none] (190) at (2.25, -1.25) {};
		\node [style=none] (191) at (0.75, -1.25) {};
		\node [style=none] (192) at (1.75, -0.25) {$A_n$};
		\node [style=none] (193) at (1.75, -1.75) {$O_n$};
		\node [style=none] (194) at (-0.75, 2.5) {};
		\node [style=none] (196) at (-0.5, 2.25) {$m$};
		\node [style=none] (197) at (-0.5, -2) {$m'$};
		\node [style=none] (198) at (-0.5, 2.5) {};
		\node [style=none] (200) at (-0.5, -2.25) {};
		\node [style=none] (201) at (-0.5, 2.5) {};
		\node [style=none] (202) at (-1.25, 2.5) {};
		\node [style=none] (203) at (-1.25, -2.25) {};
		\node [style=none] (204) at (-1.25, 0) {};
	\end{pgfonlayer}
	\begin{pgfonlayer}{edgelayer}
		\draw (68.center) to (69.center);
		\draw (69.center) to (70.center);
		\draw (70.center) to (71.center);
		\draw (71.center) to (68.center);
		\draw (79.center) to (80.center);
		\draw (79.center) to (81.center);
		\draw (79.center) to (78.center);
		\draw (82.center) to (84.center);
		\draw (84.center) to (83.center);
		\draw [bend right=90, looseness=1.75] (78.center) to (85.center);
		\draw [bend left=90, looseness=1.75] (78.center) to (85.center);
		\draw [style=arrow, bend left, looseness=0.75] (88.center) to (87.center);
		\draw [style=arrow, bend left] (89.center) to (90.center);
		\draw [style=arrow, bend left, looseness=0.75] (189.center) to (188.center);
		\draw [style=arrow, bend left] (190.center) to (191.center);
		\draw [bend left=270, looseness=1.50] (201.center) to (202.center);
		\draw [bend right=90, looseness=1.50] (203.center) to (200.center);
		\draw [style=arrow] (202.center) to (204.center);
		\draw (204.center) to (203.center);
	\end{pgfonlayer}
\end{tikzpicture} \ = \ \begin{tikzpicture}
	\begin{pgfonlayer}{nodelayer}
		\node [style=none] (68) at (-1, 2.75) {};
		\node [style=none] (69) at (0.5, 2.75) {};
		\node [style=none] (70) at (0.5, 1.5) {};
		\node [style=none] (71) at (-1, 1.5) {};
		\node [style=none] (74) at (0.25, 1.75) {};
		\node [style=none] (75) at (0.25, 2.5) {};
		\node [style=none] (76) at (0.25, 1.75) {};
		\node [style=none] (77) at (0.25, 2.5) {};
		\node [style=none] (78) at (0, 2.3) {};
		\node [style=none] (79) at (0, 2.075) {};
		\node [style=none] (80) at (0.1, 1.775) {};
		\node [style=none] (81) at (-0.1, 1.775) {};
		\node [style=none] (82) at (0.25, 2.075) {};
		\node [style=none] (83) at (-0.25, 2.075) {};
		\node [style=none] (84) at (0, 2.25) {};
		\node [style=none] (85) at (0, 2.575) {};
		\node [style=none] (86) at (-0.25, 2.075) {};
		\node [style=none] (87) at (2, 1) {};
		\node [style=none] (88) at (0.75, 1.75) {};
		\node [style=none] (89) at (1.75, 0.5) {};
		\node [style=none] (90) at (0.5, 1.25) {};
		\node [style=none] (91) at (1.5, 1.75) {$A_1$};
		\node [style=none] (92) at (0.5, 0.75) {$O_1$};
		\node [style=none] (152) at (-1, -1.25) {};
		\node [style=none] (153) at (0.5, -1.25) {};
		\node [style=none] (154) at (0.5, -2.5) {};
		\node [style=none] (155) at (-1, -2.5) {};
		\node [style=none] (156) at (0.25, -2.25) {};
		\node [style=none] (157) at (0.25, -1.5) {};
		\node [style=none] (158) at (0.25, -2.25) {};
		\node [style=none] (159) at (0.25, -1.5) {};
		\node [style=none] (160) at (0, -1.7) {};
		\node [style=none] (161) at (0, -1.925) {};
		\node [style=none] (162) at (0.1, -2.225) {};
		\node [style=none] (163) at (-0.1, -2.225) {};
		\node [style=none] (164) at (0.25, -1.925) {};
		\node [style=none] (165) at (-0.25, -1.925) {};
		\node [style=none] (166) at (0, -1.75) {};
		\node [style=none] (167) at (0, -1.425) {};
		\node [style=none] (168) at (-0.25, -1.925) {};
		\node [style=none] (179) at (0.25, -1.5) {};
		\node [style=none] (181) at (0.25, -1.5) {};
		\node [style=none] (188) at (1.75, -0.5) {};
		\node [style=none] (189) at (0.5, -1) {};
		\node [style=none] (190) at (2, -1) {};
		\node [style=none] (191) at (0.75, -1.5) {};
		\node [style=none] (192) at (1, -0.25) {$A_n$};
		\node [style=none] (193) at (1.5, -1.75) {$O_n$};
		\node [style=none] (196) at (-0.75, 2.5) {$m_1$};
		\node [style=none] (197) at (-0.75, 1.75) {$m_1'$};
		\node [style=none] (206) at (-1, -1.25) {};
		\node [style=none] (207) at (-1, -2.5) {};
		\node [style=none] (208) at (-0.75, -1.5) {$m_n$};
		\node [style=none] (209) at (-0.75, -2.25) {$m_n'$};
		\node [style=none] (214) at (-0.75, 1.5) {};
		\node [style=none] (215) at (-0.75, 2.75) {};
		\node [style=none] (216) at (-1.5, 2.75) {};
		\node [style=none] (217) at (-1.5, 1.5) {};
		\node [style=none] (218) at (-1.5, 2) {};
		\node [style=none] (219) at (-0.75, -2.5) {};
		\node [style=none] (220) at (-0.75, -1.25) {};
		\node [style=none] (221) at (-1.5, -1.25) {};
		\node [style=none] (222) at (-1.5, -2.5) {};
		\node [style=none] (223) at (-1.5, -2) {};
	\end{pgfonlayer}
	\begin{pgfonlayer}{edgelayer}
		\draw (68.center) to (69.center);
		\draw (69.center) to (70.center);
		\draw (70.center) to (71.center);
		\draw (71.center) to (68.center);
		\draw (79.center) to (80.center);
		\draw (79.center) to (81.center);
		\draw (79.center) to (78.center);
		\draw (82.center) to (84.center);
		\draw (84.center) to (83.center);
		\draw [bend right=90, looseness=1.75] (78.center) to (85.center);
		\draw [bend left=90, looseness=1.75] (78.center) to (85.center);
		\draw [style=arrow, bend left, looseness=0.75] (88.center) to (87.center);
		\draw [style=arrow, bend left] (89.center) to (90.center);
		\draw (152.center) to (153.center);
		\draw (153.center) to (154.center);
		\draw (154.center) to (155.center);
		\draw (155.center) to (152.center);
		\draw (161.center) to (162.center);
		\draw (161.center) to (163.center);
		\draw (161.center) to (160.center);
		\draw (164.center) to (166.center);
		\draw (166.center) to (165.center);
		\draw [bend right=90, looseness=1.75] (160.center) to (167.center);
		\draw [bend left=90, looseness=1.75] (160.center) to (167.center);
		\draw [style=arrow, bend left, looseness=0.75] (189.center) to (188.center);
		\draw [style=arrow, bend left] (190.center) to (191.center);
		\draw (207.center) to (206.center);
		\draw [bend left=270, looseness=1.50] (215.center) to (216.center);
		\draw [bend right=90, looseness=1.50] (217.center) to (214.center);
		\draw [style=arrow] (216.center) to (218.center);
		\draw (218.center) to (217.center);
		\draw [bend left=270, looseness=1.50] (220.center) to (221.center);
		\draw [bend right=90, looseness=1.50] (222.center) to (219.center);
		\draw [style=arrow] (221.center) to (223.center);
		\draw (223.center) to (222.center);
	\end{pgfonlayer}
\end{tikzpicture} $}
    \caption{A decentralised policy, in which the agents can be thought of as made from a collection of independent agents, each of which maintains an independent memory.}
    \label{fig:lics_pf_cartoon}
  \end{subfigure}
  \caption{ decentralised POMDPs ($\dec$-POMDPs) and decentralised agent-state policies. }
\end{figure}

Note that no constraint is placed on the behaviour of transition, observation, and reward functions, within the class of decentralised POMDPs we can however impose further constraints.
\begin{definition}[Observation independence]
\label{definition:obs_indep}
A $\dec$-POMDP is \emph{observation-independent} if for each
$i$ the component $\mathcal{O}_i(a,s)$ is independent of $a_j$ for all
$j\neq i$.
Equivalently, for all $a=(a_1,\dots,a_n)\in A$ and $s\in S$, $\mathcal{O}(a,s)=\bigl(\mathcal{O}_1(a_1,s),\dots,\mathcal{O}_n(a_n,s)\bigr)$.
\end{definition}
Observation independence is the deterministic expression of the
no-signalling constraint $A_i \not\rightarrow  \Omega_k$ for $i\neq k$ within one
environment step. In general, this kind of signalling constraint occurs quite naturally within the $\dec$-POMDP framework, that is, it is common to study environments in which the action of each agent does not influence at-that-time-step the observations of the other agents \cite{10.5555/1619332.1619356, amato}. 

In the traditional approach a policy for a decentralised POMDP consists in specifying for each agent's history of observations and actions, a new corresponding action $A_i$. 
We will consider the generalization of this picture to again consider suitable compressions of the history, with update and policy referencing an abstract memory.
\begin{definition}
We say that a deterministic agent is decentralised if $\pi = \prod_i \pi_i$ and $\mathcal{U} = \prod_i \mathcal{U}_i$, with each $\pi_i : M_i \rightarrow A_i$ and each $\mathcal{U}_i : M_i \times A_i \times \Omega_i  \rightarrow  M_i $. 
\end{definition}
In the case that one assumes the update is a concatenation of action-observation histories this recovers the traditional approach to defining decentralised policies as functions $\pi_i : \texttt{Hist}(A_i \times \Omega_i) \rightarrow A_i$.

Given an agent-state policy, we may compute the cumulative reward that it can achieve. It is assumed for agent-state policies that some memory-state is initialized $m_0$. In general one might allow the initialisation of memory to depend on some initial observation $o$ and so in turn on $s$, this case can however be incorporated by allowing for an initial "observation" round in which the assigned reward is $0$.
To compute the cumulative reward over a finite-horizon $T$ then, each action is computed as $a_t = \pi(m_t)$, the environment returns an observation $o_{t+1} = \mathcal{O}(a_t,s_t)$, updates its internal state $s_{t+1} = \mathcal{T}(a_t,s_t)$, and returns a reward, $r_{t+1} = \mathcal{R}(a_t,s_t)$. Finally, the action and observation are used to update the memory-state $m_{t+1} = \mathcal{U}(m_t, a_t, o_{t+1})$. 
\begin{definition}
    The performance of an agent-state policy given initial state distribution $\mu$ is given by $J^{(\pi, \mathcal{U})}_{\mu} =  \sum_s \mu (s) \sum_{t=1}^T \gamma^{t-1} r_{t} (m_0, s) $.
\end{definition}

\section{Process functions}
The framework of higher-order quantum operations allows one to treat quantum channels as resources and to compose them in ways that may not respect a fixed causal order \cite{chiribella_supermaps,chiribella_switch,taranto2025higherorderquantumoperations}.  In the classical deterministic limit these higher-order operations reduce to \emph{process functions}.  Such functions are of particular interest because they model deterministic closed time-like curves that remain paradox‑free \cite{Baumeler_2021}. Here we use their precise formulation given in \cite{PhysRevResearch.6.L032020}.

\begin{figure}[h]
  \centering
  \begin{subfigure}[t]{0.3\linewidth}
    \centering
    \resizebox{\linewidth}{!}{$\begin{tikzpicture}
	\begin{pgfonlayer}{nodelayer}
		\node [style=none] (9) at (-2.25, 1.75) {};
		\node [style=none] (10) at (0, 1.75) {};
		\node [style=none] (11) at (0, 1.25) {};
		\node [style=none] (12) at (-1.5, 1.25) {};
		\node [style=none] (13) at (-1.5, -1) {};
		\node [style=none] (14) at (0, -1) {};
		\node [style=none] (15) at (0, -1.5) {};
		\node [style=none] (16) at (-2.25, -1.5) {};
		\node [style=none] (17) at (-0.75, 2.5) {};
		\node [style=none] (18) at (-0.75, 1.75) {};
		\node [style=none] (19) at (-0.75, -1.5) {};
		\node [style=none] (20) at (-0.75, -2.25) {};
		\node [style=none] (21) at (-0.75, 1.25) {};
		\node [style=none] (22) at (-0.75, 0.5) {};
		\node [style=none] (23) at (-0.75, -0.25) {};
		\node [style=none] (24) at (-0.75, -1) {};
		\node [style=none] (25) at (-1.875, 1.25) {$w$};
		\node [style=none] (29) at (-0.325, 2.4) {$F$};
		\node [style=none] (31) at (-0.325, -0.65) {$I$};
		\node [style=none] (32) at (-0.325, 0.9) {$O$};
		\node [style=none] (48) at (-0.325, -2.1) {$P$};
		\node [style=none] (49) at (-1.25, 0.5) {};
		\node [style=none] (50) at (1.25, 0.5) {};
		\node [style=none] (51) at (1.25, -0.25) {};
		\node [style=none] (52) at (-1.25, -0.25) {};
		\node [style=none] (53) at (0, 0.1) {$g$};
		\node [style=none] (54) at (0.75, 0.5) {};
		\node [style=none] (55) at (0.75, -0.25) {};
		\node [style=none] (56) at (0.75, -2.25) {};
		\node [style=none] (57) at (0.75, 2.5) {};
		\node [style=none] (58) at (1.25, -2.1) {$X$};
		\node [style=none] (59) at (1.25, 2.4) {$Y$};
	\end{pgfonlayer}
	\begin{pgfonlayer}{edgelayer}
		\draw (9.center) to (10.center);
		\draw (11.center) to (12.center);
		\draw (12.center) to (13.center);
		\draw (13.center) to (14.center);
		\draw (15.center) to (16.center);
		\draw (16.center) to (9.center);
		\draw (17.center) to (18.center);
		\draw (21.center) to (22.center);
		\draw (23.center) to (24.center);
		\draw (19.center) to (20.center);
		\draw (10.center) to (11.center);
		\draw (14.center) to (15.center);
		\draw (49.center) to (50.center);
		\draw (50.center) to (51.center);
		\draw (51.center) to (52.center);
		\draw (52.center) to (49.center);
		\draw (57.center) to (54.center);
		\draw (55.center) to (56.center);
	\end{pgfonlayer}
\end{tikzpicture} \ = \  \begin{tikzpicture}
	\begin{pgfonlayer}{nodelayer}
		\node [style=none] (21) at (-0.25, 1.25) {};
		\node [style=none] (22) at (-0.25, 0.5) {};
		\node [style=none] (23) at (-0.25, -0.25) {};
		\node [style=none] (24) at (-0.25, -1) {};
		\node [style=none] (29) at (0.175, 1.15) {$F$};
		\node [style=none] (48) at (0.175, -0.85) {$P$};
		\node [style=none] (49) at (-0.5, 0.5) {};
		\node [style=none] (50) at (1, 0.5) {};
		\node [style=none] (51) at (1, -0.25) {};
		\node [style=none] (52) at (-0.5, -0.25) {};
		\node [style=none] (53) at (0.25, 0.1) {$w \star g$};
		\node [style=none] (54) at (0.75, 1.25) {};
		\node [style=none] (55) at (0.75, 0.5) {};
		\node [style=none] (56) at (1.175, 1.15) {$Y$};
		\node [style=none] (58) at (0.75, -1) {};
		\node [style=none] (59) at (0.75, -0.25) {};
		\node [style=none] (60) at (1.175, -0.9) {$X$};
	\end{pgfonlayer}
	\begin{pgfonlayer}{edgelayer}
		\draw (21.center) to (22.center);
		\draw (23.center) to (24.center);
		\draw (49.center) to (50.center);
		\draw (50.center) to (51.center);
		\draw (51.center) to (52.center);
		\draw (52.center) to (49.center);
		\draw (54.center) to (55.center);
		\draw (58.center) to (59.center);
	\end{pgfonlayer}
\end{tikzpicture}$}
    \label{fig:lics_dec_pomdp_cartoon_2}
      \caption{A process function $w$ can be evaluated on an arbitrary function $g$, using the unique fixed point criterion.}
  \end{subfigure}\hfill
  \begin{subfigure}[t]{0.48\linewidth}
    \centering
    \resizebox{\linewidth}{!}{$\begin{tikzpicture}
	\begin{pgfonlayer}{nodelayer}
		\node [style=none] (9) at (-3.25, 1.75) {};
		\node [style=none] (10) at (0.75, 1.75) {};
		\node [style=none] (11) at (0.75, 1.25) {};
		\node [style=none] (12) at (-0.75, 1.25) {};
		\node [style=none] (13) at (-0.75, -1) {};
		\node [style=none] (14) at (0.75, -1) {};
		\node [style=none] (15) at (0.75, -1.5) {};
		\node [style=none] (16) at (-3.25, -1.5) {};
		\node [style=none] (17) at (-2, 2.5) {};
		\node [style=none] (18) at (-1.25, 1.75) {};
		\node [style=none] (19) at (-1.25, -1.5) {};
		\node [style=none] (20) at (-2, -2.25) {};
		\node [style=none] (21) at (0, 1.25) {};
		\node [style=none] (22) at (0, 0.5) {};
		\node [style=none] (23) at (0, -0.25) {};
		\node [style=none] (24) at (0, -1) {};
		\node [style=none] (25) at (-1.25, 1.25) {$w$};
		\node [style=none] (29) at (-1.575, 2.4) {$F$};
		\node [style=none] (32) at (0.425, 0.9) {$O_n$};
		\node [style=none] (48) at (-1.575, -2.1) {$P$};
		\node [style=none] (49) at (-3.25, 1.25) {};
		\node [style=none] (50) at (-3.25, -1) {};
		\node [style=none] (51) at (-1.75, -1) {};
		\node [style=none] (52) at (-1.75, 1.25) {};
		\node [style=none] (53) at (-2.5, 1.25) {};
		\node [style=none] (54) at (-2.5, 0.5) {};
		\node [style=none] (55) at (-2.5, -0.25) {};
		\node [style=none] (56) at (-2.5, -1) {};
		\node [style=none] (57) at (-2.075, -0.65) {$I_1$};
		\node [style=none] (58) at (-2.075, 0.9) {$O_1$};
		\node [style=none] (59) at (-2.5, 0.5) {};
		\node [style=none] (60) at (-2.5, -0.25) {};
		\node [style=none] (66) at (0, 0.5) {};
		\node [style=none] (67) at (0, -0.25) {};
		\node [style=none] (73) at (-1.25, 0) {$\dots$};
		\node [style=none] (90) at (-2, 0.5) {};
		\node [style=none] (91) at (-4.5, 0.5) {};
		\node [style=none] (92) at (-4.5, -0.25) {};
		\node [style=none] (93) at (-2, -0.25) {};
		\node [style=none] (94) at (-3.25, 0.1) {$g_1$};
		\node [style=none] (95) at (-4, 0.5) {};
		\node [style=none] (96) at (-4, -0.25) {};
		\node [style=none] (97) at (0, -2.25) {};
		\node [style=none] (98) at (0, 2.5) {};
		\node [style=none] (99) at (0.5, -2.1) {$X_1$};
		\node [style=none] (100) at (0.5, 2.4) {$Y_1$};
		\node [style=none] (105) at (-0.5, 0.5) {};
		\node [style=none] (106) at (2, 0.5) {};
		\node [style=none] (107) at (2, -0.25) {};
		\node [style=none] (108) at (-0.5, -0.25) {};
		\node [style=none] (109) at (0.75, 0.1) {$g_n$};
		\node [style=none] (110) at (1.5, 0.5) {};
		\node [style=none] (111) at (1.5, -0.25) {};
		\node [style=none] (112) at (1.5, -2.25) {};
		\node [style=none] (113) at (1.5, 2.5) {};
		\node [style=none] (114) at (2, -2.1) {$X_n$};
		\node [style=none] (115) at (2, 2.4) {$Y_n$};
		\node [style=none] (116) at (0.425, -0.6) {$I_n$};
	\end{pgfonlayer}
	\begin{pgfonlayer}{edgelayer}
		\draw (9.center) to (10.center);
		\draw (11.center) to (12.center);
		\draw (12.center) to (13.center);
		\draw (13.center) to (14.center);
		\draw (15.center) to (16.center);
		\draw [in=90, out=-90] (17.center) to (18.center);
		\draw (21.center) to (22.center);
		\draw (23.center) to (24.center);
		\draw [in=90, out=-90] (19.center) to (20.center);
		\draw (10.center) to (11.center);
		\draw (14.center) to (15.center);
		\draw (9.center) to (49.center);
		\draw (49.center) to (52.center);
		\draw (52.center) to (51.center);
		\draw (51.center) to (50.center);
		\draw (50.center) to (16.center);
		\draw (53.center) to (54.center);
		\draw (55.center) to (56.center);
		\draw (90.center) to (91.center);
		\draw (91.center) to (92.center);
		\draw (92.center) to (93.center);
		\draw (93.center) to (90.center);
		\draw [in=90, out=-90] (98.center) to (95.center);
		\draw [in=90, out=-90] (96.center) to (97.center);
		\draw (105.center) to (106.center);
		\draw (106.center) to (107.center);
		\draw (107.center) to (108.center);
		\draw (108.center) to (105.center);
		\draw (113.center) to (110.center);
		\draw (111.center) to (112.center);
	\end{pgfonlayer}
\end{tikzpicture} \ = \  \begin{tikzpicture}
	\begin{pgfonlayer}{nodelayer}
		\node [style=none] (0) at (-0.5, 1.25) {};
		\node [style=none] (1) at (-0.5, 0.5) {};
		\node [style=none] (2) at (-0.5, -0.25) {};
		\node [style=none] (3) at (-0.5, -1) {};
		\node [style=none] (4) at (-0.075, 1.15) {$F$};
		\node [style=none] (5) at (-0.075, -0.85) {$P$};
		\node [style=none] (6) at (-0.75, 0.5) {};
		\node [style=none] (7) at (1.25, 0.5) {};
		\node [style=none] (8) at (1.25, -0.25) {};
		\node [style=none] (9) at (-0.75, -0.25) {};
		\node [style=none] (10) at (0.25, 0.1) {$w \star \{ g_i \}$};
		\node [style=none] (11) at (1, 1.25) {};
		\node [style=none] (12) at (1, 0.5) {};
		\node [style=none] (13) at (1.675, 1.15) {$\times_i Y_i$};
		\node [style=none] (14) at (1, -1) {};
		\node [style=none] (15) at (1, -0.25) {};
		\node [style=none] (16) at (1.675, -0.9) {$ \times_i X_i$};
	\end{pgfonlayer}
	\begin{pgfonlayer}{edgelayer}
		\draw (0.center) to (1.center);
		\draw (2.center) to (3.center);
		\draw (6.center) to (7.center);
		\draw (7.center) to (8.center);
		\draw (8.center) to (9.center);
		\draw (9.center) to (6.center);
		\draw (11.center) to (12.center);
		\draw (14.center) to (15.center);
	\end{pgfonlayer}
\end{tikzpicture}$}
    \label{fig:lics_dec_pomdp_cartoon_3}
      \caption{A multi-input process function can be evaluated on an arbitrary family of functions using the unique fixed point criterion.}
  \end{subfigure}
  \label{fig:lics_cartoons_2}
  \caption{Process functions and multi-input process functions.}
\end{figure}
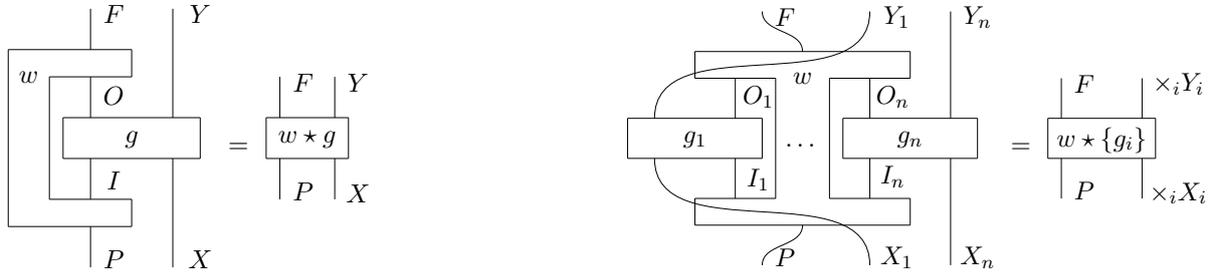

\begin{definition}[One-input process function]
Let $P$, $O$, $F$ and $I$ be sets.  A \emph{one-input process function} of type $(I\!\rightarrow  O)\rightarrow (P\!\rightarrow  F)$ is a function $w: P\times O\rightarrow  F\times I$ such that for every function $g: I\rightarrow  O$ and every $p\in P$, the equation $
o = g\bigl(w^I(p,o)\bigr),
$ has a unique solution $o\in O$.  
\end{definition}
It may not be immediately clear why such a fixed-point criterion allows for $w$ to be interpreted as a higher-order function. 
\begin{definition}
    Given any process function $w:(I\!\rightarrow  O)\rightarrow (P\!\rightarrow  F) $ we define its contraction on a function $g: I \times X \rightarrow O \times Y$ to be the function $w \star g : P \times X \rightarrow F \times Y$ where $(w\star g)(p,x) = (w^{F}(p, o)  ,  g^Y(w^I(p,o), x)    )$ and
    $o$ is the unique fixed point, that is, the unique solution to the equation $o = g^O_x \bigl(w^I(p,o)\bigr)$ with $g_x(\iota) := g(\iota,x)$.
\end{definition}
Process functions can be generalized to multiple-inputs, in which case they allow for non-causal concatenations. In the following, $w^{I_i}$ denotes the $I_i$-component of $w$, so that
$
w(p,o_1,\dots,o_n)
=
\bigl(w^F(p,\vec o),\,w^{I_1}(p,\vec o),\dots,w^{I_n}(p,\vec o)\bigr).
$
\begin{definition}[$n$-input process functions]
\label{definition:n_process_function}
An \emph{$n$-input process function} is a function
$
w: P\times O_1\times\cdots\times O_n
\rightarrow
F\times I_1\times\cdots\times I_n
$
satisfying the following unique fixed-point condition: for every choice of deterministic functions
$
g_i: I_i \rightarrow  O_i
$
and every $p\in P$, the system of equations
$
{o_{i}}
=
g_i\!\bigl(w^{I_i}(p,\vec o)\bigr)
$
has a unique solution $\vec o=(o_1,\dots,o_n)\in O$.
\end{definition}
Multi-input process functions can be evaluated on products of functions, in a simple extension of the evaluation for single-input process functions. 
\begin{definition}
    The contraction of a multi-input process function $w: P\times O_1\times\cdots\times O_n
\rightarrow
F\times I_1\times\cdots\times I_n$ on a family of functions $\{g_i: I_i \times X_i \rightarrow O_i \times Y_i \}$, is denoted $w \star \{g_i\} : P\times\prod_i X_i \rightarrow F\times\prod_i Y_i$, and defined by $w \star \{g_i\}(p, \vec x) = \Big(w^{F}(p, \vec o) , \prod_i  g^{Y_i}_i \big( \bigl(w^{I_i}(p,\vec o)\bigr) ,x_i \big)  \Big)$ where $\vec o$ is the unique fixed point for the equation $o_{i} = {g_i}^{O_i}_{{\vec x}_i}\bigl(w^{I_i}(p,\vec o)\bigr)$ (with ${g_i}_{x_i}(\iota) := g_i( \iota ,x_i)$).
\end{definition}
The primary motivation for process functions and their quantum generalisations is in the study of \textit{indefinite causal order}. Whereas definite and indefinite causal order are typically defined in term of the possibility to violate a causal inequality, we take the equivalent definition of \cite{dourdent2025paradoxfreeclassicalnoncausalityunambiguous} for ease of presentation. 
\begin{definition}
An $n$-partite process function $w : P \times ( \times_{i=1}^n O_i) \to F \times ( \times_{i=1}^n I_i )$ has a definite causal order if and only if the following two conditions hold:
\begin{enumerate}
\item[(i)] There exists at least one factor $I_k$ such that $w^{I_k}$ is independent of all $\vec o_{-k} \in\prod_{i \neq k} O_i$.
\item[(ii)] For every party $k$ and for every $o_{k}\in O_{k}$, the contraction $ w_{o_k}^{( \times_{i=1}^n I_i) \backslash I_k}: P \times ( \times_{i=1}^n O_i)  \backslash O_k  \rightarrow F \times   ( \times_{i=1}^n I_i ) \backslash I_k $ satisfies $(i)$, where $w_{o}$ is contraction of $w$ with the constant function $f(\iota) = o$. 
\end{enumerate}
\end{definition}
There exists, indeed, multi-input process functions for which it cannot be said which agent from $1 \dots n$ comes first, because, the choice of each agent is permitted to signal to the outcome of any other agent. 
An important example of such a process function is the \textit{Lugano process} $w_{\texttt{Lugano}} : \{\bullet\} \times O_1\times\cdots\times O_n
\rightarrow
 \{\bullet\} \times I_1\times\cdots\times I_n$ \cite{Baumeler_2016, AraujoFeix_PrivateComm_2014} which can be compactly defined by $
\iota_1 := o_3\,(o_2 \oplus 1)$,
$\iota_2 := o_1\,(o_3 \oplus 1)$, and 
$\iota_3 := o_2\,(o_1 \oplus 1) $ \cite{dourdent2025paradoxfreeclassicalnoncausalityunambiguous}.

\section{Correspondence between agents and process functions}
To relate agent-state policies to process functions we begin by formalising a single round of interaction between a policy and a POMDP.  Let $\mathcal{A}=(\pi,\mathcal{U})$ be a deterministic agent-state policy with memory set $M$ and let $\mathcal{P}=\langle S,A,\Omega,T,O,R\rangle$ be a deterministic POMDP.  Given an initial memory state $m\in M$ and environment state $s\in S$, the one‑step interaction proceeds as follows:
\begin{enumerate}
  \item The agent chooses an action $a=\pi(m)$,
  \item The environment:
  \begin{itemize}
      \item updates its state to $s'=\mathcal{T}(a,s)$,
      \item outputs an observation $o=\mathcal{O}(a,s)$,
      \item produces a reward $r=\mathcal{R}(a,s)$,
  \end{itemize}
  \item The agent updates its memory according to $m' = \mathcal{U}(m,a,o)$.
\end{enumerate}
We write $(\pi,\mathcal{U})\bullet \langle \mathcal{T},\mathcal{O},\mathcal{R} \rangle: M\times S\rightarrow  M\times S\times \realsgeq$ for the function mapping $(m,s)$ to $(m',s',r)$ defined by following this procedure. 
\begin{definition}
Let $(\pi_{\mathcal{A}},\mathcal{U}_{\mathcal{A}})$ and $(\pi_{\mathcal{B}},\mathcal{U}_{\mathcal{B}})$ be two deterministic agent-state policies with the same memory space.  We say they are behaviourally \emph{equivalent}, written $(\pi_{\mathcal{A}},\mathcal{U}_{\mathcal{A}}) \cong (\pi_{\mathcal{B}},\mathcal{U}_{\mathcal{B}})$, if for every deterministic POMDP $\langle \mathcal{T},\mathcal{O},\mathcal{R} \rangle$ and every memory-state/environment-state pair $(m,s)$ then $((\pi_{\mathcal{A}},\mathcal{U}_{\mathcal{A}})\bullet \langle \mathcal{T},\mathcal{O},\mathcal{R} \rangle)(m,s) \,=\,((\pi_{\mathcal{B}},\mathcal{U}_{\mathcal{B}})\bullet \langle \mathcal{T},\mathcal{O},\mathcal{R} \rangle )(m,s)$.  
\end{definition}
Concretely, behavioural equivalence requires that:
\begin{itemize}
    \item $\mathcal{T}\bigl(\pi_{\mathcal{A}}(m),s\bigr)=\mathcal{T}\bigl(\pi_{\mathcal{B}}(m),s\bigr)$,
    \item $
\mathcal{R}\bigl(\pi_{\mathcal{A}}(m),s\bigr)=\mathcal{R}\bigl(\pi_{\mathcal{B}}(m),s\bigr)$,
    \item $\mathcal{U}_{\mathcal{A}}\bigl(m,\pi_{\mathcal{A}}(m),\mathcal{O}(\pi_{\mathcal{A}}(m),s)\bigr)=\mathcal{U}_{\mathcal{B}}\bigl(m,\pi_{\mathcal{B}}(m),\mathcal{O}(\pi_{\mathcal{B}}(m),s)\bigr)$.
\end{itemize}
This definition captures the idea that agent-state policies can use different updating mechanisms and still induce the same behavior and so collect the same cumulative reward.  

We now prove that equivalence classes of such agents are in bijection with one‑input process functions.
To do so, let us first establish a basic proposition regarding $1$-input process functions, that they exhibit a comb-decomposition generalising that of quantum and stochastic higher order processes \cite{chiribella_supermaps, Baumeler_2016, kissinger_caus}.
\begin{proposition}
Let $w: P\times O\rightarrow  F\times I$ be a one-input process function.  Then there exists functions $w^F: P\times O\rightarrow  F$ and $w^I: P\rightarrow  I$ such that for all $p\in P$ and $o\in O$ then $
w(p,o) = \bigl(w^F(p,o),\,w^I(p)\bigr)
$.
\end{proposition}
\begin{proof}
Write $w(p,o)=(w^F(p,o),w^I(p,o))$.  Suppose for contradiction that there exist $o,o'\in O$ with $w^I(p,o)\neq w^I(p,o')$ for some fixed $p\in P$.  Define a function $f: I\rightarrow  O$ by setting $f\bigl(w^I(p,o)\bigr)=o$ and $f\bigl(w^I(p,o')\bigr)=o'$.  The fixed‑point equation $\bar{o}=f\bigl(w^I(p,\bar{o})\bigr)$ then has two solutions $\bar{o}=o$ and $\bar{o}=o'$, contradicting the unique fixed‑point condition.  Therefore $w^I(p,o)$ must be constant as a function of $o$, and we may write $w(p,o)=\bigl(w^F(p,o),\,w^I(p)\bigr)$.
\end{proof}
We note, that a deterministic POMDP $\langle \mathcal{T},\mathcal{O},\mathcal{R} \rangle$ can be packaged into a single function $\mathcal{P}_{\langle \mathcal{T},\mathcal{O},\mathcal{R} \rangle}: A \times S \rightarrow \Omega \times S \times \realsgeq$, where $\mathcal{P}^{\Omega} = \mathcal{O}$, $\mathcal{P}^{S} = \mathcal{T}$, $\mathcal{P}^{\realsgeq} = \mathcal{R}$.
\newlength{\imgH}
\setlength{\imgH}{6cm}

\begin{theorem}
There is a one‑to‑one correspondence between equivalence classes of deterministic agent-state policies and one‑input process functions of type $(A \rightarrow \Omega)\rightarrow (M \rightarrow  M)$, such that for each equivalence class $[\pi, \mathcal{U}]$ the associated process function $w_{[\pi, \mathcal{U}]}$ satisfies $[\pi, \mathcal{U}] \bullet \langle \mathcal{T},\mathcal{O},\mathcal{R} \rangle = w_{[\pi, \mathcal{U}]} \star \mathcal{P}_{ \langle \mathcal{T},\mathcal{O},\mathcal{R} \rangle }$.
\end{theorem}
\begin{proof}
A complete proof is given in the appendix, in the main text we give the constructions witnessing the equivalence, along with the correspondence between process-function contraction and $1$-step evaluation. 
Fix a deterministic agent $(\pi,\mathcal{U})$ with memory space $M$, action set $A$ and observation set $\Omega$.  We construct a function
$
w_{[\pi, \mathcal{U}]}: M\times\Omega\longrightarrow M\times A$, by defining
$w_{[\pi, \mathcal{U}]}(m,o) := \bigl(\mathcal{U}(m,\pi(m),o),\,\pi(m)\bigr)$.
Conversely, let $w: M\times\Omega\rightarrow  M\times A$ be a one‑input process function.  By the decomposition of process functions, there are functions $w^M: M\times\Omega\rightarrow  M$ and $w^A: M\rightarrow  A$ such that $w(m,o)=(w^M(m,o),w^A(m))$.  We define a deterministic agent-state policy $(\pi_w, \mathcal{U}_w)$ with $\pi_w:=w^A$ and $\mathcal{U}_w(m,a,o):=w^M(m,o)$.  
Importantly, the sequence of constructions (agent-state policy) $\to$ (process function) $\to$ (agent-state policy), returns a \textit{different} policy, but, one which is \textit{behaviourally equivalent}. 

Finally, recall that $(w\star \mathcal{P})(m,s) = (w^{M}(m, o)  ,  \mathcal{P}^{S \times \realsgeq}(w^A(m,o), s))$ where $o$ is the unique solution to the equation $o = \mathcal{P}^{\Omega} \bigl((w^A(m,o)\bigr) ,s) = \mathcal{P}^{\Omega} \bigl((w^A(m)\bigr) ,s)$, so that indeed:
\begin{align*}
(w\star \mathcal{P})(m,s) & = (\mathcal{U}(m,\pi(m), \mathcal{P}^{\Omega}(w^A(m),s)) , \mathcal{P}^{S \times \realsgeq}(w^A(m),s))  \\ 
& =  (\mathcal{U}(m,\pi(m), \mathcal{O}(w^A(m),s)) , \mathcal{T}(w^A(m),s),  \mathcal{R}(w^A(m),s)) \\
 &  =  (\mathcal{U}(m,\pi(m), \mathcal{O}(\pi(m),s)) , \mathcal{T}(\pi(m),s),  \mathcal{R}(\pi(m),s))  \\
 &  = [\pi, \mathcal{U}] \bullet \langle \mathcal{T},\mathcal{O},\mathcal{R} \rangle (m,s) ,
\end{align*}
and so, the $1$-step evaluation is given by application of the corresponding process function as a higher-order function. 
\end{proof}

This theorem can be extended to the decentralised case, in order to do so we identify the additional condition on one-party process functions which allows them to be interpreted as multiple non-communicating-between-rounds parties. 
\begin{definition}
    We say that a function $w : \prod_{i=1}^n M_i \times  \prod_{i=1}^n \Omega_i \longrightarrow \prod_{i=1}^n M_i \times \prod_{i=1}^n A_i$ is between-rounds decentralised if it decomposes as $w(m,\vec o) = \big(\prod_{i=1}^n w^{M_i}(m_i,{o_{i}}) ,   \prod_{i=1}^n w^{A_i}(m_i, \vec o)\big).$
\end{definition}
\begin{corollary}
    The equivalence between process functions and agent-state policies, restricts to an equivalence between between-rounds decentralised process functions and decentralised agent-state policies.
\end{corollary}
\begin{proof}
    Note that if a decentralised function is also a one-input process function, then, it further satisfies
\[
w(\vec m, \vec o) = (\prod_{i=1}^n w^{M_i}(m_i,{o_{i}}) ,   \prod_{i=1}^n w^{A_i}(m_i)),
\]
and so we find that $\mathcal{U}_w (\vec m, \vec a, \vec o) = \prod_i w^{M_i}(m_i,{o_{i}}) = \prod_i \mathcal{U}_i(m_i,a_i,{o_{i}})$ where each $\mathcal{U}_i(m_i,a_i,{o_{i}})$ is given by $w^{M_i}(m_i,{o_{i}})$. Similarly, we find that $\pi_w (\vec  m) =  \prod_{i=1}^n w^{A_i}(m_i) = \prod_{i=1}^n \pi_i(m_i)$ where each $\pi_i := w^{A_i}(m_i)$.
\end{proof}
\section{Composing multi-agent policies and POMDPs}
In this section we use the identification between agent-state policies and process functions, to build composition rules for multi-agent policies and decentralised environments. In order to do so, we generalize process functions to act on arbitrary types. 
We identify types as those sets which are closed under fixed-point-duals, in analogy with the normaliser-duals of \cite{kissinger_caus}.
\begin{definition}
For any subset $\mathscr{P}\subseteq \mathbf{Set}(A,A')$ the dual set $\mathscr{P}^{*} \subseteq \mathbf{Set}(A',A)$ is defined by $\mathscr{P}^{*} := \{ w \in \mathbf{Set}(A',A) \ s.t \ \forall \mathcal{P} \in \mathscr{P} \ \exists ! \sigma :  \ \mathcal{P}(w(\sigma)) = \sigma  \}$.
\end{definition}
The move to select types via a duality relation is, more broadly, in the spirit of the double-glueing/orthogonality constructions used to build models of linear logic \cite{HYLAND2003183,Schalk_2015}.
\begin{definition}
A \textit{type} is any subset $\mathscr{P} \subseteq \mathbf{Set}(A,A')$ such that $\mathscr{P}^{**} = \mathscr{P}$.
\end{definition}
\begin{proposition}
    Types are closed under duals and products, more precisely:
    \begin{itemize}
        \item If $\mathscr{P}$ is a type then $\mathscr{P}^*$ is a type,
    \item If $\mathscr{P}, \mathscr{Q}$ are types then the product $\mathscr{P} \times \mathscr{Q} = \{ (\mathcal{P}, \mathcal{Q}) \vert \mathcal{P} \in \mathscr{P} \wedge \mathcal{Q}  \in \mathscr{Q}   \}$ is a type.
    \end{itemize}
\end{proposition}
\begin{proof}
Given in Appendix B. 
\end{proof}
\begin{definition}
    A process function $w: \mathscr{P} \rightarrow \mathscr{Q}$ between sets $\mathscr{P} \subseteq  \mathbf{Set}(A,A')$ and $\mathscr{Q} \subseteq  \mathbf{Set}(B,B')$ is a function $ w: B \times A' \rightarrow B' \times A $ such that for every $\mathcal{P} \in \mathscr{P}$ and $b \in B$ then the fixed point equation $ \mathcal{P}(w^{A}(b, \sigma)) = \sigma, $ has a unique solution $\sigma(b)$ where furthermore the fixed point contraction of $w$ with $p$ denoted $ w_p : B \rightarrow B' $ and defined by $ w_p(b) = w^{B'}(b,\sigma(b))$, is an element of $\mathscr{Q}$.
\end{definition}
The definition of morphisms is reminiscent of the \(\mathrm{Int}\) construction, which freely embeds a traced symmetric monoidal category into a compact closed one \cite{JoyalStreetVerity1996Traced}. In the present setting, however, the relevant trace is only partially defined (in the sense of partially traced categories) \cite{Malherbe_2012}. All-together then, the following category $\mathbf{PF}$ can likely be understood as the result applying a double glueing procedure to the result of the partial-version of the \(\mathrm{Int}\) construction.
\begin{theorem}
Process functions form a category $\mathbf{PF}$. More precisely, given process functions $w: \mathscr{R} \rightarrow \mathscr{Q}$ and $u: \mathscr{Q} \rightarrow \mathscr{P}$ with $\mathscr{P} \subseteq \mathbf{Set}(A,A')$, $\mathscr{Q} \subseteq \mathbf{Set}(B,B')$, and $\mathscr{R} \subseteq \mathbf{Set}(C,C')$:
\begin{itemize}
\item there exists a unique solution $\tau$ to the equation $ \tau = u^B (a, w^B (\tau , c' ))$,
\item the function $u \star w : \mathscr{R} \rightarrow \mathscr{P}$ defined by 
    $ u \star w (a,c') = (u^A (a, w^B (\tau , c' )) , w^C (\tau , c')), $
    is a process function $\mathscr{R} \rightarrow \mathscr{P}$,
    \item contraction is associative $(u \star w) \star v = u \star  (w \star v)$,
\end{itemize}
and furthermore, defining $1_{\mathscr{P}} = \texttt{swap}_{AA'} : A \times A' \rightarrow A' \times A $, then:
\begin{itemize}
\item contraction is unital $u \star 1_\mathscr{Q} = w = 1_\mathscr{P} \star w$ .
\end{itemize}
\end{theorem}
\begin{proof}
    Given in Appendix B.
\end{proof}
\begin{theorem}
The category $\mathbf{PF}$ is furthermore $*$-autonomous with
    \begin{itemize}
        \item Dualizing operator given by $(-)^{*}$,
        \item Tensor given by the product $\mathscr{P} \otimes \mathscr{Q} =  \mathscr{P} \times \mathscr{Q}$,
        \item Par given as de-Morgan dual to the product $\mathscr{P} \parr \mathscr{Q} = (\mathscr{P}^* \times \mathscr{Q}^*)^{*}$.
    \end{itemize}
\end{theorem}
\begin{proof}
    Given in Appendix B.
\end{proof}
We now show how to identify the compositional rules of star-autonomous categories with concepts in the study of decentralised multi-agent systems. Let us note that the set of constant functions $\texttt{Const}(A,A') \subset \mathbf{Set}(A,A')$ is a type for any pair $A,A'$, and that $\texttt{Const}(A,A')^* = \mathbf{Set}(A,A')$.
As a result note that any singleton type $\{ \bullet \}$ must be of the form  $\texttt{Const}(\{ \bullet' \},A)$ for some $A$ and so sets of the form $\mathbf{Set}(\{\bullet\} , A)$ can be uniquely characterised as the duals of singleton types. 
In keeping with the language of \cite{kissinger_caus} we refer to duals of singletons as first-order systems. This notion is justified by the realisation that for any sets $A,B$ we can identify the functions $A \rightarrow B$ with the process functions $\mathbf{Set}(\{\bullet\} , A) \rightarrow \mathbf{Set}(\{\bullet\} , B)$. Finally, we use the symbol $\Rightarrow$ to represent the closure operator, more precisely, $A \Rightarrow B := A^* \parr B$. 
\begin{theorem}
    Taking all objects $A,B,A',B', \dots$ to be first order, the following identifications can be made;
    \begin{itemize}
        \item $A \Rightarrow A' \cong \mathbf{Set}(A,A')$,
        \item $(A \Rightarrow A')  \otimes (B \Rightarrow B')  \cong \mathbf{Set}(A , A' ) \times \mathbf{Set}(B,B')$,
        \item $(A \Rightarrow A')  \parr (B \Rightarrow B')  \cong \mathbf{Set}(A \times B , A' \times B' ) $.
        \end{itemize}
        Consequently, 
        \begin{itemize}
        \item $ (A \Rightarrow \Omega) \parr (S \Rightarrow (S \times \realsgeq)) \cong $ POMDPs,
        \item $ (\parr_i (A_i \Rightarrow \Omega_i)) \parr (S \Rightarrow (S \times \realsgeq)) \cong $ $\dec$-POMDPs,
        \item $ (\otimes_i (A_i \Rightarrow \Omega_i)) \parr (S \Rightarrow (S \times \realsgeq)) \cong $ observation-independent $\dec$-POMDPs.
    \end{itemize}
    Furthermore,
    \begin{itemize}
        \item $ (A \Rightarrow \Omega) \Rightarrow (M \Rightarrow M') \cong $ process functions,
        \item $   \otimes_i ((A_i \Rightarrow A_i') \Rightarrow (B_i \Rightarrow B_i')) \cong $ decentralised process functions,
                \item $ ( \otimes_i (A_i \Rightarrow \Omega_i') ) \Rightarrow (M \Rightarrow M') \cong $ multi-input process functions.
    \end{itemize}
\end{theorem}
\begin{proof}
Given in Appendix B.
\end{proof}
There is a common sub-expression ($\otimes_i (A_i \Rightarrow \Omega_i)$) which appears both within observation independence and within indefinite causal orders (multi-input process functions). This commonality allows for the construction via the cut, of a consistent notion of cumulative reward for any indefinite order strategy over any observation-independent $\dec$-POMDP. 
\begin{definition}
    The evaluation of a process function of the form $w :  \mathscr{Q} \rightarrow (M \Rightarrow M)$ on a POMDP of the form $\mathcal{P} \in \mathscr{Q} \parr (S \Rightarrow (S \times \realsgeq))$ is defined by the cut along $\mathscr{Q}$, that is, $(w \parr 1_{S \Rightarrow (S \times \realsgeq)})\circ \mathcal{P}$.
\end{definition}
The result of this cut has type $ (M \Rightarrow M) \parr (S \Rightarrow (S \times \realsgeq)) \cong  (M \times S) \Rightarrow ( M \times S \times \realsgeq)$, for ease of notation we will denote this function constructed from the $1$-step evaluation as
$(w\star \mathcal{P}) \cong (w \parr 1_{S \Rightarrow (S \times \realsgeq)})\circ \mathcal{P}$ which generalises the usual contraction of $1$-input process functions.
The result of the $t$-step interaction can then be expressed in terms of $(w\star \mathcal{P})$ by
\begin{itemize}
\item $F_1  := (w\star \mathcal P) : M\times S \to M\times S\times \realsgeq$, 
\item $F_{t+1}   := (f \otimes 1_{\realsgeq^{t}})\circ F_t$, 
\end{itemize}
where $F_t : M\times S \to M\times S\times \realsgeq^{t}$.
\begin{definition}
For each $\mathcal{P} \in \mathscr{Q} \parr (S \Rightarrow (S \times \realsgeq))$ and $w : \mathscr{Q} \rightarrow (M \Rightarrow M)$, along with discount factor $\gamma\in[0,1)$, distribution $\mu$ on $S$, time-horizon $T$, and initial memory state $m_0$, we define the process function performance  \[ J^{w}_{\mu} := \sum_{s_0\in S}\mu(s_0) \sum_{t=1}^{T}\gamma^{t-1}\, r_t(m_0, s_0), \]
where $r_t$ is computed from the last component of $(\bigcirc_{k = 1}^{t}( (w \star \mathcal{P}) \otimes  1_{\otimes_{i = 1}^k \realsgeq})(m_0,s_0)$, i.e., $r_t = (\bigcirc_{k = 1}^{t}( (w \star \mathcal{P}) \otimes  1_{\otimes_{i = 1}^k \realsgeq})^{t+2}(m_0,s_0)$.
\end{definition}
\section{Indefinite Causal Order Advantage in a Decentralised POMDP}
We now focus on the observation-independent decentralised case in which $\mathscr{Q} = \otimes_i (A_i \Rightarrow \Omega_i)$, and identify a separation between the performance of between-rounds decentralised process functions with and without indefinite causal order. To do so we introduce a causal game, and a construction from games to $\dec$-POMDPs.
\begin{definition}
\label{def:maj-gyni}
For $x=(x_1,x_2,x_3)\in\{ 0,1 \}^3$, define the majority bit
\[
\mathsf{maj}(x) :=
\begin{cases}
1 & \text{if } x_1+x_2+x_3 \ge 2,\\
0 & \text{otherwise.}
\end{cases}
\]
Define $\mathcal{Q}:\{ 0,1 \}^3\times \{ 0,1 \}^3\to\{0,1\}$ by $\mathcal{Q}(x,y)=1$ iff either
\begin{align*}
& \mathsf{maj}(x)=0 \ \text{ and }\ \forall i\in\{1,2,3\}: y_{[i+1]_3}=x_i \\
\textrm{or } & \mathsf{maj}(x)=1 \ \text{ and }\ \forall i\in\{1,2,3\}: y_i=1\oplus x_{[i+1]_3}.
\end{align*}
We refer to this as the \emph{majority--GYNI} game.
\end{definition}
\begin{definition}
\label{def:gyni-decpomdp}
Fix integers $T\ge 1$ (horizon) and $n\ge 0$ (number of warm-up rounds), and a
discount factor $\gamma\in(0,1]$, we define the deterministic $\dec$-POMDP $\mathcal{P}^{(n)}_{\mathrm{GYNI}}$ with:
\begin{itemize}
\item global state space $S := \{ 0,1 \}^3 \times \{0,1,\dots,T\}$ with state $(x,k)$,
\item local action spaces $A_i:=\{ 0,1 \}$, so $A=\{ 0,1 \}^3$ with joint action $y=(y_1,y_2,y_3)$,
\item local observation spaces $\Omega_i:=\{ 0,1 \}$, so $\Omega=\{ 0,1 \}^3$,
\item observation function (observation-independent): $
\mathcal{O}_i(y,(x,k)) := x_i $,
\item transition: $x$ is constant and the counter increments: $\mathcal{T}(y,(x,k)) := (x,k+1) $,
\item reward:
\[
\mathcal{R}(y,(x,k)) :=
\begin{cases}
0 & \text{if } k<n,\\
\mathcal{Q}(x,y) & \text{if } k\ge n,
\end{cases}
\]
\item Initial distribution: uniform over all states such that $k=0$.
\end{itemize}
\end{definition}
In other words, the first $n$ rounds return reward $0$, and the game is played between rounds $n+1,\dots,T$.
\begin{lemma}
\label{lem:local-memory-invariant}
In $\mathcal{P}^{(n)}_{\mathrm{GYNI}}$, for each agent $i$ and each round $t$ there exists
a function $\varphi_{i,t}:\{ 0,1 \}\to M_i$ such that $m_i^t = \varphi_{i,t}(x_i)$.
\end{lemma}
In other words, the variable $m_i^t$ is independent of $x_j$ for $j\neq i$.
\begin{proof}
We prove the statement by induction on $t$. At $t=0$, $m_i^0$ is fixed hence
$m_i^0=\varphi_{i,0}(x_i)$ for the constant function $\varphi_{i,0}(x) = m_i^0 \ (\forall x)$.
Now, let us assume that $m_i^t=\varphi_{i,t}(x_i)$.
By decentralisation, its memory update at the end of the round uses only $(m_i^t, x_i)$.
Formally, we have that:
    \begin{align*}
        m_i^{t+1} & = (w\star \mathcal{P})^{M_i}(\prod_i m_i^t, s_t) \\
         \texttt{process function contraction}  & = w^{M_i}(\prod_i  \varphi_{i,t}(x_i), x) \\
        \texttt{between-rounds decentralisation} & = w^{M_i}(\prod_i  \varphi_{i,t}(x_i), x_i) \\
         & = \varphi_{i,t+1}(x_i),
    \end{align*}
    thereby completing the induction step. 
\end{proof}
\begin{lemma}
\label{lem:gybi-3over4}
Consider a fixed $i \in\{1,2,3\}$, a function $g : X_i \rightarrow Y_i$, and let $x$ be uniformly distributed.
Then for any possible choice of the remaining two output bits $y_{-i}$ we have that,
\[
\Pr\bigl[\mathcal{Q}(x,y)=1 \bigm|   y_i=g(x_i) \bigr] \le \frac{3}{4}.
\]
\end{lemma}
\begin{proof}
Without loss of generality we take $i=1$.
Fix $b\in\{ 0,1 \}$ and restrict to the four inputs with $x_1=b$. Inspecting the definition of the GYNI game (equivalently, the truth table for $\mathcal{Q}$ given in Appendix C), among these four inputs the winning value of $y_1$ equals $1$ for exactly one input and equals $0$ for the other three. Therefore, once $y_1$ is fixed to $g(b)$, at least one of the four cases with $x_1=b$ must lose, so the conditional success probability given $x_1=b$ is at most $3/4$. Averaging over each of the two possibilities for $b$ completes the proof.
\end{proof}
\begin{theorem}
\label{thm:definite-order-bound}
Consider $\mathcal{P}^{(n)}_{\mathrm{GYNI}}$ with uniform initial $x$ and any between-rounds decentralised multi-input process function which has a definite causal order. Then, its return satisfies
\[
J \;\le\; \frac{3}{4}\sum_{t=n+1}^{T}\gamma^{t-1}.
\]
\end{theorem}

\begin{proof}
Fix any reward round $t\ge n+1$. By definite causality there exists a party $i^\star$ such that $y_{i^\star}^t$ depends only on $(m_{i^\star}^t,x_{i^\star})$. Furthermore, by between-rounds decentralisation, $m_{i^\star}^t$ is itself a function only of $x_{i^\star}$, so $y_{i^\star}^t=g(x_{i^\star})$ for some $g:\{ 0,1 \}\to\{ 0,1 \}$. Consequently, on each round we have that
\(
\Pr[\mathcal{Q}(x,y^t)=1]\le 3/4
\),
hence $\mathbb{E}[r_t]\le 3/4$ for each $t\ge n+1$.
Since $r_t=0$ for $t\le n$, we obtain
\[
J=\sum_{t=1}^{T}\gamma^{t-1}\mathbb{E}[r_t]
=\sum_{t=n+1}^{T}\gamma^{t-1}\mathbb{E}[r_t]
\le \frac{3}{4}\sum_{t=n+1}^{T}\gamma^{t-1}.
\]
\end{proof}
\begin{theorem}[Perfect performance with indefinite causal order]
\label{thm:indefinite-order-achieves}
There exists a between-rounds decentralised multi-input process function (with indefinite causal order) such that in every reward round $t\ge n+1$,
the realised output $y^t$ satisfies $\mathcal{Q}(x,y^t)=1$.
Consequently,
\[
J \;=\; \sum_{t=n+1}^{T}\gamma^{t-1}.
\]
\end{theorem}
\begin{proof}
The \emph{Lugano} process $w_{\texttt{Lugano}}$, is known to win the majority-vote GYNI game perfectly \cite{Baumeler_2016,AraujoFeix_PrivateComm_2014,dourdent2025paradoxfreeclassicalnoncausalityunambiguous}. Concretely, one may define the multi-input process function $\overline{w}_{\texttt{Lugano}} : (\prod_i M_i ) \times (\prod_i X_i) \rightarrow (\prod_i M_i )  \times (\prod_i Y_i)$ with trivial singleton-set memories $M_i = \{ \bullet \}$, and with $\overline{w}_{\texttt{Lugano}}^{A_i} := w_{\texttt{Lugano}}^{A_i} $.
Then, $r_t=\mathcal{Q}(x,y^t)=1$ for all $t\ge n+1$, while $r_t=0$ for $t\le n$ by construction of
$\mathcal{P}^{(n)}_{\mathrm{GYNI}}$.
Substituting into the definition for process function performance gives the claimed value.
\end{proof}

Consequently, for any $T\ge n+1$ and any $\gamma\in(0,1]$ we obtain a strict gap between the process function performances
achievable with and without the assumption of a definite background causal structure.
\section{Conclusion}

The correspondence established in this article shows that deterministic agents in artificial intelligence (up-to behavioural equivalence) and one‑input process functions in the foundations of physics are mathematically equivalent. In other words, two agents are behaviourally indistinguishable precisely when they induce the same process function.  This correspondence comes with a duality of interpretations: in physics $w$ is viewed as the spacetime environment into which local operations are inserted, whereas in artificial intelligence $w$ encodes the agent’s decision‑making procedure and the inserted map corresponds to the environment. 


Regarding future work, with the connection made, precise methods developed for higher‑order causality could now be transported to decision‑making problems.  First, is the question of whether there exist practical already known observation-independent decentralised-POMDPs \cite{OliehoekAmato2016decPOMDP}, in which general multi-input process function strategies outperform traditional definite-ordered ones. Furthermore, it is unclear at this stage how efficiently indefinite causal order strategies can be constructed or learned, via for instance a suitable generalization of policy iteration to process-function iteration. Natural open questions in this direction regard the computational complexity of finding the optimal agent-state policies for different $\textit{types}$ of POMDPs.

Furthermore, it is natural to wonder whether types can be endowed with the connectives of a spatio-temporal logic, such as BV-logic \cite{Guglielmi_2007}, (noting that stochastic and quantum higher-order processes form BV-categories \cite{simmons2022higherordercausaltheoriesmodels}), possibly allowing for the purely logical/compositional representation of communication (and constrained communication) within decentralised multi-agent systems \cite{OliehoekAmato2016decPOMDP}.

Regarding quantum computation and the possibility of quantum artificial intelligence, the results of this paper motivate a particular fully quantum generalization of POMDPs. More precisely, the upgrading of POMDPs from functions to quantum channels of type $\mathcal{P}: A\times S\longrightarrow \Omega\times S\times R$ with each $A,S,\Omega,S,R$ a Hilbert space (with the greatest complication coming from defining a suitable Hilbert space $R$ for coherently collecting rewards). A quantum decision-making agent in this perspective then corresponds to a (possibly multi-input) quantum super-channel (process matrix). Exactly how this viewpoint compares with the quantum partially observable Markov decision processes of \cite{PhysRevA.90.032311}, the quantum Markov decision processes of \cite{saldi2024quantummarkovdecisionprocesses},
as well as quantum games \cite{Gutoski_2007}, quantum agents for algorithmic discovery \cite{kerenidis2025quantumagentsalgorithmicdiscovery}, and potential advantages of quantum resources in multi-agent protocols \cite{anantharam2025quantumadvantagedecentralizedcontrol}, is again left for future work.

\bibliography{bibliography}

\appendix

\section{Equivalence between agent-state policies and process functions}

\begin{theorem}
There is a one‑to‑one correspondence between equivalence classes of deterministic agents and one‑input process functions of type $(A \rightarrow \Omega)\rightarrow (M \rightarrow  M)$, such that for each equivalence class $[\pi, \mathcal{U}]$ the associated process function $w_{[\pi, \mathcal{U}]}$ satisfies $[\pi, \mathcal{U}] \bullet \langle \mathcal{T},\mathcal{O},\mathcal{R} \rangle = w_{[\pi, \mathcal{U}]} \star \mathcal{P}_{\langle \mathcal{T},\mathcal{O},\mathcal{R}_{\geq} \rangle}$.
\end{theorem}
\begin{proof}
\emph{(Agent-state policies $\Rightarrow$ process functions).}  Fix a deterministic agent $\mathcal{A}=(\pi,\mathcal{U})$ with memory space $M$, action set $A$ and observation set $\Omega$.  We construct a function
$
w_{[\pi, \mathcal{U}]_{\cong}}: M\times\Omega\longrightarrow M\times A$, by defining
$w_{[\pi, \mathcal{U}]}(m,o) := \bigl(\mathcal{U}(m,\pi(m),o),\,\pi(m)\bigr)$.
To check that $w_{[\pi, \mathcal{U}]}$ is a one‑input process function note that for any function $f: A\rightarrow \Omega$ and any $m\in M$ the fixed‑point equation $o = f\bigl(w_{[\pi, \mathcal{U}]}^I(m,o)\bigr)$ becomes $o = f\bigl(\pi(m)\bigr)$, which has a unique solution since the right‑hand side is independent of $o$.  

Now suppose $[\pi_{\mathcal{A}}, \mathcal{U}_{\mathcal{A}}] \cong [\pi_{\mathcal{B}}, \mathcal{U}_{\mathcal{B}}]$.  We show that $w_{[\pi_{\mathcal{A}}, \mathcal{U}_{\mathcal{A}}]}=w_{[\pi_{\mathcal{B}}, \mathcal{U}_{\mathcal{B}}]}$.  Fix $m\in M$ and $o\in\Omega$.  Consider the POMDP with state space $S=\Omega\times A$, transition function $\mathcal{T}(a,(o,a'))= (o',a)$ for some fixed state $o'$, observation function $\mathcal{O}(a,(o,a'))=o$, and reward function trivial $\mathcal{R}(a,(o,a)) = 1$.  For the single step of interaction with starting state $(m,(o,a'))$, since $[\pi_{\mathcal{A}}, \mathcal{U}_{\mathcal{A}}] \cong [\pi_{\mathcal{B}}, \mathcal{U}_{\mathcal{B}}]$ we have that
$
([\pi_{\mathcal{A}}, \mathcal{U}_{\mathcal{A}}] \bullet \langle \mathcal{T},\mathcal{O},\mathcal{R} \rangle) (m,(o,a')) = ([\pi_{\mathcal{B}}, \mathcal{U}_{\mathcal{B}}]\bullet \langle \mathcal{T},\mathcal{O},\mathcal{R} \rangle) (m,(o,a')),
$
and therefore $
\mathcal{U}_{A}\bigl(m,\pi_{A}(m),o\bigr)=\mathcal{U}_{B}\bigl(m,\pi_{B}(m),o\bigr)$ and $\pi_{A}(m)=\pi_{B}(m)$.  It follows that $w_{[\pi_{\mathcal{A}}, \mathcal{U}_{\mathcal{A}}]}(m,o)=w_{[\pi_{\mathcal{B}}, \mathcal{U}_{\mathcal{B}}]}(m,o)$ for all $m,o$, and hence equivalent agent-state policies induce the same process function.

\emph{(Process functions $\Rightarrow$ agent-state policies).} Conversely, let $w: M\times\Omega\rightarrow  M\times A$ be a one‑input process function.  By the decomposition of process functions, there are functions $w^F: M\times\Omega\rightarrow  M$ and $w^I: M\rightarrow  A$ such that $w(m,o)=(w^F(m,o),w^I(m))$.  We define a deterministic agent-state policy $(\pi_w, \mathcal{U}_w)$ with $\pi_w:=w^I$ and $\mathcal{U}_w(m,a,o):=w^F(m,o)$.  

We now show that these two constructions are inverses of each-other up to behavioral equivalence.  Starting with a process function $w'$, constructing the agent $(\pi_{w'},\mathcal{U}_{w'})$ and then forming the associated process function $w_{(\pi_{w'},\mathcal{U}_{w'})}$ gives
\begin{align*}
w_{(\pi_{w'},\mathcal{U}_{w'})}(m,o)  &=  \bigl(\mathcal{U}_{w'}(m,\pi_{w'}(m),o),\,\pi_{w'}(m)\bigr)  \\
 &=  \bigl({w'}^F(m,o),\,{w'}^I(m)\bigr) \\
 &=  {w'}(m,o),
\end{align*}
and so $w_{(\pi_{w'},\mathcal{U}_{w'})}={w'}$.  Conversely, start with an agent-state policy $(\pi,\mathcal{U})$, form $w_{(\pi',\mathcal{U}')}$ and then construct the agent-state policy $(\pi_{w_{(\pi',\mathcal{U}')}},\mathcal{U}_{w_{(\pi',\mathcal{U}')}})$.  One easily checks that $\pi_{w_{(\pi',\mathcal{U}')}}=\pi$ and then we have that
\begin{align*}
 \mathcal{U}_{w_{(\pi',\mathcal{U}')}} (m, \pi_{{(\pi',\mathcal{U}')}}(m) , \mathcal{O}(\pi_{{(\pi',\mathcal{U}')}}(m), s)) 
& =   w^F_{(\pi',\mathcal{U}')} (m , \mathcal{O}(\pi'(m), s))    \\
& =     U^{'} (m, \pi'(m) , \mathcal{O}(\pi'(m), s)),
\end{align*} 
and similarly for rewards, so that $(\pi_{w_{(\pi',\mathcal{U}')}},\mathcal{U}_{w_{(\pi',\mathcal{U}')}}) \cong (\pi', \mathcal{U}')$.  This shows that the constructions are mutually inverse up-to equivalence and so establishes a bijection between equivalence classes of agent-state policies and one‑input process functions.

Finally, recall that $(w\star \mathcal{P})(m,s) = (w^{M}(m, o)  ,  \mathcal{P}^{S \times \realsgeq}(w^A(m,o), s))$ where $o$ is the unique solution to the equation $o = \mathcal{P}^{\Omega} \bigl((w^A(m,o)\bigr) ,s) = \mathcal{P}^{\Omega} \bigl((w^A(m)\bigr) ,s)$, so that indeed:
\begin{align*}
(w\star \mathcal{P})(m,s) & = (\mathcal{U}(m,\pi(m), \mathcal{P}^{\Omega}(w^A(m),s)) , \mathcal{P}^{S \times \realsgeq}(w^A(m),s))  \\ 
& =  (\mathcal{U}(m,\pi(m), \mathcal{O}(w^A(m),s)) , \mathcal{T}(w^A(m),s),  \mathcal{R}(w^A(m),s)) \\
 &  =  (\mathcal{U}(m,\pi(m), \mathcal{O}(\pi(m),s)) , \mathcal{T}(\pi(m),s),  \mathcal{R}(\pi(m),s))  \\
 &  = [\pi, \mathcal{U}] \bullet \langle \mathcal{T},\mathcal{O},\mathcal{R} \rangle (m,s).
\end{align*}
This completes the proof. 
\end{proof}

\section{Composition of POMDP types}

\begin{proposition}\label{lem:residual-exists}\label{lem:residual-exists_1}
Taking the following definitions for upper and lower duals
\begin{itemize}
\item Upper dual: $\mathscr{P}^{*} = \{ w \in \mathbf{Set}(A',A) \ s.t \ \forall \mathcal{P} \in \mathscr{P} \ \exists ! \sigma \ \mathcal{P}(w(\sigma)) = \sigma  \}$,
\item Lower dual: $\mathscr{P}_{*} =  \{ w \in \mathbf{Set}(A',A) \ s.t \ \forall \mathcal{P} \in \mathscr{P} \ \exists ! \rho \ w(\mathcal{P}(\rho) )= \rho  \}$,
\end{itemize}
then $\mathscr{P}^{*} = \mathscr{P}_{*}$.
\end{proposition}
\begin{proof}
Consider $w:A' \rightarrow A$ and $p$ in $\mathscr{P}$, then consider any solution to the equation $w(\mathcal{P}(\rho)) = \rho$. Let us take $\sigma = \mathcal{P}(\rho)$, then we see that $w(\sigma) = \rho$ and so indeed $\sigma = \mathcal{P}(\rho) = \mathcal{P}(w(\sigma))$. 
Conversely given any solution to the equation $\mathcal{P}(w(\sigma)) = \sigma$. Let us take $\rho = w(\sigma)$, then we see that $\mathcal{P}(\rho) = \sigma$ and so indeed $\rho = w(\sigma) = w(\mathcal{P}(\rho))$.
Note that given two distinct solutions $ w(\mathcal{P}(\rho_1) = \rho_1 \neq  \rho_2 =  w(\mathcal{P}(\rho_2)) \implies \mathcal{P}(\rho_1) \neq \mathcal{P}(\rho_2)$, and so if there are two distinct solutions for the former equation there are two for the latter and vice-versa.
Consequently, if there is a unique solution to the former equation there is a solution to the latter, furthermore there cannot be two solutions to the latter, else there would be two solutions to the former, therefore there is a unique solution to the former equation if and only if there is a unique solution to the latter equation. 
\end{proof}
\begin{proposition}
    Types are closed under duals and products, more precisely:
    \begin{itemize}
        \item If $\mathscr{P}$ is a type then $\mathscr{P}^*$ is a type,
    \item If $\mathscr{P}, \mathscr{Q}$ are types then the product $\mathscr{P} \times \mathscr{Q} = \{ (\mathcal{P}, \mathcal{Q}) \vert \mathcal{P} \in \mathscr{P} \wedge  \mathcal{Q}  \in \mathscr{Q}   \}$ is a type.
    \end{itemize}
\end{proposition}
\begin{proof}
It is easy to see that for any set $\mathscr{P}$ then $\mathscr{P}^*$ is a \textit{type}. Indeed, using the fact that for any set $\mathscr{M}$ then $\mathscr{M} \subseteq \mathscr{M}^{**}$, and that for any pair of sets $\mathscr{N} \subseteq \mathscr{M}$ then $\mathscr{M}^{*} \subseteq \mathscr{N}^{*}$ we have that $\mathscr{P}^{*} \subseteq \mathscr{P}^{***}$ and $\mathscr{P}^{***} \subseteq \mathscr{P}^{*}$ so that $\mathscr{P}^{*} = \mathscr{P}^{***}$.

Regarding the product, consider that for any $\mathcal{S} \in (\mathscr{P} \times \mathscr{Q})^{**}$ we have that for every $w \in (\mathscr{P} \times \mathscr{Q})^{*}$ then the equation $w(\mathcal{S}(\sigma)) = \sigma$ has a unique fixed point $\sigma_w$. We can construct an important class of the $w  \in (\mathscr{P} \times \mathscr{Q})^{*}$ as those which exhibit arbitrary communication from $A$ to $B'$, that is, consider any $a_o$ and any $f: A' \rightarrow B$ then construct $w(a',b') := (a_0, f(a'))$. This function is indeed in  $(\mathscr{P} \times \mathscr{Q})^{*}$ since for any pair $\mathcal{P},\mathcal{Q} \in \mathscr{P} \times \mathscr{Q}$ then the equation $(a',b') = (\mathcal{P} \times \mathcal{Q}) (w(a',b')) = (\mathcal{P}(a_0) , \mathcal{Q}(f(a')))$ has a unique solution $a'  = \mathcal{P}(a_0) , \ b' = \mathcal{Q}(f(\mathcal{P}(a_0)))$. Using this class of $w$, the fixed point equation for $\mathcal{S}$ reduces to $ (\sigma_A, \sigma_B) = w(\mathcal{S}(\sigma_A, \sigma_B)) = (a_0 , f(\mathcal{S}^A(\sigma_A, \sigma_B)))$ which has solutions given by taking $\sigma_A = a_0$ and taking $\sigma_B$ to solve $\sigma_B = f(\mathcal{S}^A(a_0, \sigma_B))$. Consequently, the unique fixed point condition for $\mathcal{S}$ entails that for each $a_0$ and $f$ there is a unique solution to the equation $\sigma_B = f(\mathcal{S}^A(a_0, \sigma_B))$, this in turn entails by the decomposition proposition for process functions that $\mathcal{S}^A$ is independent to $b$. Symmetric reasoning gives that $\mathcal{S}^B$ is independent of $a$ and so in full we have that $\mathcal{S}(a,b) = (\mathcal{S}^A(a), \mathcal{S}^B(b))$. 

Finally, we check that $\mathcal{S}^A,\mathcal{S}^B$ are in $\mathscr{P}^{**}, \mathscr{Q}^{**}$ respectively. Indeed consider a pair $(w_p,w_q) \in \mathscr{P}^{*} \times \mathscr{Q}^{*}$ then note that for any $(\mathcal{P}, \mathcal{Q}) \in \mathscr{P}^{*} \times \mathscr{Q}^{*}$ the fixed point equation $ (\mathcal{P} \times \mathcal{Q} )(w_p \times w_q)(a', b') = (a',b')$ has a unique solution given by the unique solutions to $ \mathcal{P}(w_p(a')) = a'$ and  $ \mathcal{Q}(w_q(b')) = b'$ respectively, consequently, $w_p \times w_q \in (\mathscr{P} \times \mathscr{Q})^{*}$. We then note that for any $\mathcal{S} = \mathcal{S}^A \times \mathcal{S}^B \in (\mathscr{P} \times \mathscr{Q})^{**}$ there is by definition a unique solution to the fixed point equation $ (w_p\times w_q)(\mathcal{S}^A \times \mathcal{S}^B)(a, b) = (a,b)$ and so a unique solution to the pair of fixed point equations $ w_p(\mathcal{S}^A(a)) = a$ and $ w_q(\mathcal{S}^B(b)) = b$ respectively, this confirms that $\mathcal{S}^A \in \mathscr{P}^{**}$ and similarly that $\mathcal{S}^B \in \mathscr{Q}^{**}$, which completes the proof.
\end{proof}

\begin{proposition}\label{lem:residual-exists_2}
The par $\parr$ and dual $(-)^*$ can be used to construct the space of all process functions, more precisely:
$h \in \mathscr P\parr\mathscr Q \iff h : \mathscr Q^* \rightarrow \mathscr P$.
\end{proposition}
\begin{proof}
We first confirm that for every $a\in A$ the equation
$ b' = h_{B'}\bigl(a,\pi(b')\bigr)$ has a unique solution.
Fix $a\in A$ and $\pi \in \mathscr Q^*$, and recall that the constant map $\kappa_a:A'\to A$ lies in $\mathscr{P}^*$.
As a result we have that for  $(\kappa_a,\pi)\in\mathscr{P}^*\times\mathscr Q^*$, and $h\in(\mathscr{P}^*\times\mathscr Q^*)^*$, the fixed-point equation
\[
  (\kappa_a\times \pi)\bigl(h(x,y)\bigr)=(x,y)
\]
has a unique solution $(x,y)\in A\times B$.
Reading off the first component gives $x=\kappa_a(h_{A'}(x,y))=a$, and so for each $a$ there exists a unique solution to the equation $b' = h_{B'}\bigl(a,\pi(b')\bigr)$ induced by the second component.
Now, consider the pair $(w,\pi)\in\mathscr{P}^*\times\mathscr Q^*$ and again $h\in(\mathscr{P}^*\times\mathscr Q^*)^*$, note that the fixed-point equation
\begin{equation}\label{eq:fp-w-pi}
  (w\times\pi)\bigl(h(a,b)\bigr)=(a,b)
\end{equation}
has a unique solution $(a,b)\in A\times B$.
Let $b' =h_{B'}(a,b)$, then  $b=\pi(b')$ and $b'=h_{B'}(a,\pi(b'))$.
As a result, $h_{\pi}(a)=h_{A'}(a,\pi(b'))$.
Now, the first component of the fixed point equation for $w,\pi,h$ entails that for all $w,a$ then $a=w(h_{A'}(a,b))=w(h_{\pi}(a))$ has a unique fixed-point solution, so that indeed $h_{\pi} \in \mathscr{P}^{**} = \mathscr{P}$.
What remains is to check the reverse inclusion, that $\{ \mathscr{P} \rightarrow \mathscr{Q} \} \subseteq \mathscr{Q} \parr \mathscr{P}^{*}$. Indeed consider some $w \in \{ \mathscr{P} \rightarrow \mathscr{Q} \}$ and consider the equation \[(e_q \times p) \circ w (b,a') = (b,a')    \] with $\mathcal{P} \in \mathscr{P}$ and $e_\mathcal{Q} \in \mathscr{Q}^{*}$. Recall that since $w: \mathscr{P} \rightarrow \mathscr{Q}$ then for each $\mathcal{P} \in \mathscr{P}$ we can compute the residual $w_p \in \mathscr{Q}$, consequently let us solve the equation by setting $b$ to be the unique solution to $e_q  ( w_p (b)) = b$ and then setting $a'$ to be the unique solution to $a' = p ( w (b,a')) =a'$. This assignment indeed solves the equation $(e_q \times p) \circ w (b,a') = (b,a')$, and since any solution must satisfy those same equations, the solution is unique. Consequently, we have proven that $\{ \mathscr{P} \rightarrow \mathscr{Q} \} \subseteq (\mathscr{Q}^{*} \times \mathscr{P})^{*} = (\mathscr{Q}^{*} \times \mathscr{P}^{**})^{*} = \mathscr{Q} \parr \mathscr{P}^{*}$.

\end{proof}

\begin{proposition}
Process functions form a category $\mathbf{PF}$. More precisely, given process functions $w: \mathscr{R} \rightarrow \mathscr{Q}$ and $u: \mathscr{Q} \rightarrow \mathscr{P}$ with $\mathscr{P} \subseteq \mathbf{Set}(A,A')$, $\mathscr{Q} \subseteq \mathbf{Set}(B,B')$, and $\mathscr{R} \subseteq \mathbf{Set}(C,C')$:
\begin{itemize}
\item there exists a unique solution $\tau$ to the equation $ \tau = u^B (a, w^B (\tau , c' ))$,
\item the function $u \star w : \mathscr{R} \rightarrow \mathscr{P}$ defined by 
    $ u \star w (a,c') = (u^A (a, w^B (\tau , c' )) , w^C (\tau , c')), $
    is a process function $\mathscr{R} \rightarrow \mathscr{P}$
\end{itemize}
and using this definition for composition of process functions along with defining $1_{\mathscr{P}} = \texttt{swap}_{AA'} : A \times A' \rightarrow A' \times A $, then:
\begin{itemize}
\item contraction is associative $(u \star w) \star v = u \star  (w \star v)$,
\item contraction is unital $u \star 1_\mathscr{Q} = w = 1_\mathscr{P} \star w$ 
\end{itemize}
\end{proposition}
\begin{proof}
       We consider process functions $w: \mathscr{R} \rightarrow \mathscr{Q}$ and $u: \mathscr{Q} \rightarrow \mathscr{P}$ with $\mathscr{P} \subseteq \mathbf{Set}(A,A')$, $\mathscr{Q} \subseteq \mathbf{Set}(B,B')$, and $\mathscr{R} \subseteq \mathbf{Set}(C,C')$. We construct the process function $u \star w : \mathscr{R} \rightarrow \mathscr{P}$ as 
    \[ u \star w (a,c') = (u^A (a, w^B (\tau , c' )) , w^C (\tau , c')), \]
    where $\tau$ is the unique solution to the equation \[ \tau = u^B (a, w^B (\tau , c' )). \]
To see the existence of the unique solution to the equation note that $w^B (\tau , c' ) = w_{f_{c'}} (\tau)$ where $f_{c'}$ is the constant function $f_{c'}(c) = c'$ and further note that since the constant functions are contained within every type, then $f_{c'} \in \mathscr{R}$ and so $ w_{f_{c'}} \in \mathscr{Q}$. As a result, the fixed point equation for $\tau = u^B (a, w^B (\tau , c' ))$ reads as $\tau = u^B (a, w_{f_{c'}} (\tau))$ where by definition $ w_{f_{c'}} \in \mathscr{Q}$ and $u: \mathscr{Q} \rightarrow \mathscr{P}$, and so there must exist a unique solution. 

    To confirm that this contraction $u \star w$ returns a process function $ \mathscr{R} \rightarrow \mathscr{P}$ we must first check the unique fixed point criterion against all $r \in \mathscr{R}$, meaning we check solutions of $ (u \star w)^C (a,r(c)) = c $, which amounts to solving 
\begin{align}
      c & = w^C (\tau , r(c)) \\ 
     \tau & = u^B (a, w^B (\tau , r(c) )).
\end{align}
This set of equations indeed admits a solution, by setting $\tau$ to be the unique solution to the equation $\tau = u^B (a, w_r(\tau))$ and taking $c$ to be the unique solution to the equation $c = w^{C}(\tau , r(c))$, then we confirm that the former equation holds vacuously and for the latter we have that $u^B (a, w^B (\tau , r(c) )) = u^B (a, w_r (\tau ))$. Since each equation individually admits only one solution, there can be no more than one simultaneous solution to the pair of equations. 

Next we must check that the resulting function $(u \star w)_r(\bullet) \in \mathscr{P}$.  We check this by proving that $(u \star w)_r(\bullet) = \mathcal{U}_{w_r}(\bullet)$ where since by definition for typed process functions, we have $w_r \in \mathscr{Q}$ and so $\mathcal{U}_{w_r} \in \mathscr{P}$. Indeed, note that
\begin{align}
    (u \star w)_r(a) &=  (u \star w)^{A'}(a,  r(c)) 
     \ \textrm{where} \  c =  (u \star w)^{C}(a,  r(c)) \\
    & =  u^{A'}(a,  w^{B}(\tau , r(c))) 
     \ \textrm{where} \  c =  w^{C}(\tau,  r(c))  \textrm{ and } \tau = u^{B}(a, w_r(\tau)) \\
     & =  u^{A'}(a, w_r (\tau))
\end{align}
Now see that
\begin{align}
    \mathcal{U}_{w_r}(a) =  u^{A'}(a, w_r (\tau))
     \ \textrm{where} \  \tau =   u^{B}(a, w_r(\tau))
\end{align}
and so $ (u \star w)_r(a) =  \mathcal{U}_{w_r}(a)$. 

We next check associativity for the contraction of process functions, for the left components we have that for all $r \in  \mathscr{R}$ then $
    ((u \star w) \star v)_r  =  (u \star w)_{v_r}
     =  \mathcal{U}_{w_{v_r}} 
     =  \mathcal{U}_{(w\star v)_r} 
     =  (u \star (w\star v))_r  $
from which we have that $ ((u \star w) \star v)^{A'}  = (u \star (w\star v))^{A'}  $.
For the right components, we have that
\begin{align*}
    (u \star (w \star v))^{D}(a,d) & =  (w\star v)^{D}(\tau, d)  \
    \textrm{where} \  \tau   =  u^{B}(a,  (w\star v)^{B}(\tau , d) ) \\
    & =  v^{D}(\sigma, d)  \ 
    \textrm{where} \ \sigma   =  w^{C}(\tau,  v^{C}(\sigma , d) )   \textrm{ and } 
    \tau   =  u^{B}(a,  (w\star v)^{B}(\tau , d) )
\end{align*}
noting that $(w\star v)^{B}(\tau , d) ) = w^{B}(\tau , v^{C}(\epsilon ,d))$ where $\epsilon = w^{C}(\tau , v^{C}(\epsilon , d))$ implies that $\epsilon = \sigma$ and we find that  
\begin{align*}
    (u \star (w \star v))^{D}(a,d) & =   v^{D}(\sigma, d)  \ 
    \textrm{where} \ \sigma   =  w^{C}(\tau,  v^{C}(\sigma , d) )   \textrm{ and } 
    \tau   =  u^{B}(a,   w^{B}(\tau , v^{C}(\sigma ,d)) )
\end{align*}
and that 
\begin{align}
    ((u \star w) \star v)^{D}(a,d) & =  v^{D}(\sigma, d)  \ 
    \textrm{where} \ \sigma   =  (u \star w)^{C}(a,  v^{C}(\sigma , d) ) ,
\end{align}
so where  $ \sigma   =  w^{C}(\tau,  v^{C}(\sigma , d) )$
and $  \tau   =  u^{B}(a,  w^{B}(\tau , v^{C}(\sigma , d)) ) $.
Consequently, we have that  $   (u \star (w \star v))^{D}(a,d)  = ((u \star w) \star v)^{D}(a,d)  $.
It is easy to check that the unit equations $1 \star w = w = w \star 1$ hold, indeed for the left unit we have that 
\begin{align}
    (1 \star w)(a,b') = (i^{A'}(a, w^{A'}(\tau, b')), w^{B}(\tau , b'))
\end{align}
where $\tau = i^{A}(a, w^{A'}(\tau , b')) \implies \tau = a$ and so 
\begin{align}
    (i^{A'}(a, w^{A'}(\tau, b')), w^{B}(\tau , b'))&  = (w^{A'}(a,b') , w^{B}(a,b')) \\
     & = w(a,b').
\end{align}
The right-unit equation follows similarly. 
\end{proof}
In order to prove coherence of these $\alpha$ and $\lambda$ isomorphisms, we will recover them as the images of a functor from products of bijections $\mathbf{Bij} \times \mathbf{Bij} $ to $\mathbf{PF}$.  Whenever the context is clear we assume that $\mathscr P\subseteq \mathbf{Set}(A,A')$.
\begin{definition}
\label{def:transport_type}
Let $\mathscr P\subseteq \mathbf{Set}(A,A')$ be a type and let $f:A\rightarrow B$ and
$g:A'\rightarrow B'$ be bijections, we define the \emph{transport} of $\mathscr P$ along $(f,g)$ by
\[
(f,g)\mathscr P = \{\, g\circ \mathcal{P}\circ f^{-1}\mid \mathcal{P}\in\mathscr P\,\}\ \subseteq\ \mathbf{Set}(B,B').
\]
\end{definition}

\begin{lemma}
\label{lem:transport_preserves_types}
If $\mathscr P$ is a type and $f,g$ are bijections, then $(f,g)\mathscr P$ is a type.
\end{lemma}

\begin{proof}
Recall that $h \in ((f,g)\mathscr{P})^*$ if and only if for every $gpf^{-1}$ then the equation $hgpf^{-1}(x) = x$ has a unique fixed-point $x$, which is true if and only if for each $p$ the equation $f^{-1}hg\mathcal{P}(y) = y$ has a unique fixed-point $y$. Consequently $((f,g)\mathscr{P})^* = (g^{-1},f^{-1})\mathscr{P}^*$. Using this, compatibility between the dualizer and transport we find that $((f,g)\mathscr{P})^{**} = ((g^{-1},f^{-1})\mathscr{P}^*)^* = (f,g)\mathscr{P}^{**} =  (f,g)\mathscr{P}$.

\end{proof}

\begin{lemma}
Let $\mathscr P\subseteq \mathbf{Set}(A,A')$ be a type, $f:A\rightarrow B$, $g:A'\rightarrow B'$ be bijections, and the function $\mathrm{Sh}_{f,g}$ be defined pointwise by $\mathrm{Sh}_{f,g}(b,a') := \bigl(g(a'),\,f^{-1}(b)\bigr)$:
\begin{itemize}
\item For every $\mathcal{P}\in\mathscr P$, the contraction of $\mathrm{Sh}_{f,g}$ with $p$ is $(\mathrm{Sh}_{f,g})_p = g\circ \mathcal{P}\circ f^{-1}$,
\item $\mathrm{Sh}_{f,g}$ is a process function of type $\mathscr P \rightarrow (f,g)\mathscr P$.
\end{itemize}
\end{lemma}

\begin{proof}
Fix $\mathcal{P}\in\mathscr P$ and $b\in B$. The fixed point equation for $a'\in A'$ is $a' = \mathcal{P}\bigl(\mathrm{Sh}_{f,g}^A(b,a')\bigr)=\mathcal{P}(f^{-1}(b))$,
whose right-hand side is independent of $a'$, so the solution is unique. Denoting this solution by $a'_*(b):=\mathcal{P}(f^{-1}(b))$, we compute the contraction $(\mathrm{Sh}_{f,g})_\mathcal{P}(b)=\mathrm{Sh}_{f,g}^{B'}(b,a'_*(b))=g\bigl(\mathcal{P}(f^{-1}(b))\bigr)=(gpf^{-1})(b)$, which lies in $(f,g)\mathscr P$ by definition.
\end{proof}

\begin{proposition}
The assignment $\mathrm{Sh}_{f,g}$ defines a functor $\mathrm{Sh}_{-,=}: \mathbf{Bij} \times \mathbf{Bij} \rightarrow  \mathbf{PF}$, concretely, we have that:
\begin{itemize}
\item $\mathrm{Sh}_{f_2,g_2}\ \star\ \mathrm{Sh}_{f_1,g_1} =
\mathrm{Sh}_{f_2\circ f_1,\;g_2\circ g_1}$,
\item $\mathrm{Sh}_{\mathrm{id}_A,\mathrm{id}_{A'}} = 1_{\mathscr P}=\texttt{swap}_{A,A'}$.
\end{itemize}
\end{proposition}

\begin{proof}
We begin with preservation of composition, evaluating both sides at $(c,a')\in C\times A'$. In $\mathrm{Sh}_{f_2,g_2}\star \mathrm{Sh}_{f_1,g_1}$, First, we find the unique fixed point $b\in B$ which satisfies $b=(\mathrm{Sh}_{f_2,g_2})^{B}(c,\mathrm{Sh}_{f_1,g_1}^{B'}(b,a')) = f_2^{-1}(c)$. Substituting into the definition of contraction then gives $(\mathrm{Sh}_{f_2,g_2}\star \mathrm{Sh}_{f_1,g_1})(c,a') = \bigl(g_2(g_1(a')),\ f_1^{-1}(f_2^{-1}(c))\bigr) = \mathrm{Sh}_{f_2\circ f_1,\;g_2\circ g_1}(c,a')$.
The identity preservation claim is immediate.
\end{proof}
Note, that as a consequence $\mathrm{Sh}_{f,g}$ is an isomorphism with inverse $\mathrm{Sh}_{f^{-1},g^{-1}}$.
Furthermore, note that transport commutes with products, meaning that $(f\times f',\,g\times g')(\mathscr P\times \mathscr Q) = (f,g)\mathscr P \times (f',g')\mathscr Q$.
\begin{lemma}
\label{lem:shuffle_tensor}
The functor $\mathrm{Sh}_{-,=}$ is homomorphic from $\times_{\mathbf{Bij}}$ to $\times_{\mathbf{PF}}$, more precisely;
\[
\mathrm{Sh}_{f,g}\times_{\mathbf{PF}}\mathrm{Sh}_{f',g'} =
\mathrm{Sh}_{f \times_{\mathbf{Bij}} f',\;g \times_{\mathbf{Bij}} g'}.
\]
\end{lemma}

\begin{proof}
Immediate by definition. 
\end{proof}

Let us now fix the standard coherence bijections in $\mathbf{Set}$ for the cartesian product as follows:
\begin{align*}
\alpha^{\mathbf{Set}}_{A,B,C} & : (A\times B)\times C \rightarrow A\times(B\times C),\\
\lambda^{\mathbf{Set}}_A & : 1\times A\rightarrow A,\\
\rho^{\mathbf{Set}}_A & : A\times 1\rightarrow A,\\
\sigma^{\mathbf{Set}}_{A,B} & :A\times B\rightarrow B\times A.
\end{align*}

\begin{definition}
\label{def:structural_as_shuffle}
Given types $\mathscr P\subseteq\mathbf{Set}(A,A')$, $\mathscr Q\subseteq\mathbf{Set}(B,B')$, and
$\mathscr R\subseteq\mathbf{Set}(C,C')$, we define:
\begin{itemize}
\item $\alpha^{\otimes}_{\mathscr P,\mathscr Q,\mathscr R} :=
\mathrm{Sh}_{\alpha^{\mathbf{Set}}_{A,B,C},\ \alpha^{\mathbf{Set}}_{A',B',C'}} : (\mathscr P\times\mathscr Q)\times\mathscr R \rightarrow \mathscr P\times(\mathscr Q\times\mathscr R)$,
\item $\lambda^{\otimes}_{\mathscr P} :=\mathrm{Sh}_{\lambda^{\mathbf{Set}}_{A},\ \lambda^{\mathbf{Set}}_{A'}}: I\times\mathscr P\to \mathscr P$,
\item $\rho^{\otimes}_{\mathscr P} :=\mathrm{Sh}_{\rho^{\mathbf{Set}}_{A},\ \rho^{\mathbf{Set}}_{A'}}: \mathscr P\times I\to \mathscr P$,
\item $\sigma_{\mathscr P,\mathscr Q} :=\mathrm{Sh}_{\sigma^{\mathbf{Set}}_{A,B},\ \sigma^{\mathbf{Set}}_{A',B'}}: \mathscr P\times\mathscr Q \to \mathscr Q\times\mathscr P$.
\end{itemize}
\end{definition}

\begin{theorem}
    The category $\mathbf{PF}$ of process functions is $*$-autonomous  with 
    \begin{itemize}
        \item Dualizing operator given by $(-)^{*}$,
        \item Tensor given by the product $\mathscr{P} \otimes \mathscr{Q} =  \mathscr{P} \times \mathscr{Q}$,
        \item Par given as de-Morgan dual to the product $\mathscr{P} \parr \mathscr{Q} = (\mathscr{P}^* \times \mathscr{Q}^*)^{*}$.
    \end{itemize}
\end{theorem}
\begin{proof}
We first confirm the monoidal structure. 
We define the action of the product on morphisms component-wise simple by taking the product of $w: \mathscr{P}_1 \rightarrow \mathscr{P}_2$ and $u: \mathscr{Q}_1 \rightarrow \mathscr{Q}_2$ with $\mathscr{P}_i \subseteq \mathbf{Set}(A_i , A_i')$ and $\mathscr{Q}_i \subseteq \mathbf{Set}(B_i , B_i')$  to be $(w \times_{\mathbf{PF}} u) (a_2 , b_2 , a_1' , b_1' )  = (w^{A_2'}(a_2,a_1') , u^{B_2'}(b_2,b_1'), w^{A_1}(a_2,a_1'), u^{B_1}(b_2,b_1'))$ which indeed has type  $ \mathscr{P}_1 \times \mathscr{Q}_1 \rightarrow \mathscr{P}_2 \times \mathscr{Q}_2$ since given any element $(\mathcal{P}_1,\mathcal{Q}_1)$ of $\mathscr{P}_1 \times \mathscr{Q}_1$ we have that the equation 
\[  (w^{A_1}(a_2, \mathcal{P}_1(a_1) ) , u^{B_1}(b_2, \mathcal{Q}_1(b_1) )) = (a_1, b_1) , \]
has unique solutions computed component-wise, and furthermore we have that by construction, component-wise, for each $(\mathcal{P}_1, \mathcal{Q}_1) \in \mathscr{P}_1 \times \mathscr{Q}_2$, then $w_{\mathcal{P}_1} \times \mathcal{U}_{\mathcal{Q}_1} \in \mathscr{P}_2 \times \mathscr{Q}_2$. 
For the symmetry we introduce the process function $\texttt{SWAP}_{\mathbf{PF}} : \mathscr{P} \times \mathscr{Q} \rightarrow \mathscr{Q} \times \mathscr{P} $ which as a function has type $\texttt{SWAP}_{\mathbf{PF}}  = (B \times A) \times (A' \times B') \rightarrow (B' \times A') \times (A \times B)$ and is defined by $\texttt{SWAP}_{\mathbf{PF}} = \texttt{swap}_{(A' \times B')  ,(B \times A) } \circ  ( \texttt{swap}_{B',A'} \times \texttt{swap}_{A,B})$.

Regarding the interchange law, consider process functions $w:\mathscr{P}_1\to \mathscr P_2$, $\overline w:\mathscr P_2\to\mathscr P_3 $, 
$u:\mathscr{Q}_1\to\mathscr Q_2$, $ \overline u:\mathscr Q_2\to\mathscr Q_3$, which as functions have types $ w: A_2\times A_1'\to A_2'\times A_1$, $ \overline w: A_3\times A_2'\to A_3'\times A_2$, $u: B_2\times B_1'\to B_2'\times B_1$, $ \overline u: B_3\times B_2'\to B_3'\times B_2$. We now check $(\overline w\times_{\mathbf{PF}}\overline u)\star (w\times_{\mathbf{PF}}u)$ on an element $(a_3,b_3,a_1',b_1')$. Recall there there is a unique fixed point $\tau=(\tau_A,\tau_B)\in A_2\times B_2$ for the equation
\begin{equation}\label{eq:fp-interchange}
(\tau_A,\tau_B) = (\overline w\times_{\mathbf{PF}}\overline u)^{A_2\times B_2}\!\Bigl( (a_3,b_3), (w\times_{\mathbf{PF}}u)^{A_2'\times B_2'}\bigl((\tau_A,\tau_B),(a_1',b_1')\bigr) \Bigr).
\end{equation}
We now recall that \[ (w\times_{\mathbf{PF}}u)^{A_2'\times B_2'}(\tau_A,\tau_B,a_1',b_1') = \bigl(w^{A_2'}(\tau_A,a_1'),\,u^{B_2'}(\tau_B,b_1')\bigr), \]
and \[ (\overline w\times_{\mathbf{PF}}\overline u)^{A_2\times B_2}(a_3,b_3,a_2',b_2') = \bigl(\overline w^{A_2}(a_3,a_2'),\,\overline u^{B_2}(b_3,b_2')\bigr). \] Using these expressions for $\times_{\mathbf{PF}}$ we have that \[ \tau_A=\overline w^{A_2}\bigl(a_3,\,w^{A_2'}(\tau_A,a_1')\bigr), \qquad \tau_B=\overline u^{B_2}\bigl(b_3,\,u^{B_2'}(\tau_B,b_1')\bigr). \]
Consequently, $\tau_A$ is the fixed point used to define the composite $\overline w\star w$, and $\tau_B$ is the fixed point used to define $\overline u\star u$. Having computed the relevant fixed point $\tau$ we further have that;
\begin{align*}
&\bigl((\overline w\times_{\mathbf{PF}}\overline u)\star (w\times_{\mathbf{PF}}u)\bigr)(a_3,b_3,a_1',b_1') \\
&=
\Bigl( (\overline w\times_{\mathbf{PF}}\overline u)^{A_3'\times B_3'}\bigl((a_3,b_3), (w\times_{\mathbf{PF}}u)^{A_2'\times B_2'}((\tau_A,\tau_B),(a_1',b_1'))\bigr),
\, \\ 
&\qquad (w\times_{\mathbf{PF}}u)^{A_1\times B_1}\bigl((\tau_A,\tau_B),(a_1',b_1')\bigr) \Bigr) \\
&= \Bigl( \overline w^{A_3'}\bigl(a_3,\,w^{A_2'}(\tau_A,a_1')\bigr), \overline u^{B_3'}\bigl(b_3,\,u^{B_2'}(\tau_B,b_1')\bigr), \, \\
&\qquad w^{A_1}(\tau_A,a_1'), u^{B_1}(\tau_B,b_1') \Bigr) \\
&= \Bigl( (\overline w\star w)^{A_3'}(a_3,a_1'), (\overline u\star u)^{B_3'}(b_3,b_1'), \, \\
&\qquad (\overline w\star w)^{A_1}(a_3,a_1'), (\overline u\star u)^{B_1}(b_3,b_1') \Bigr) \\
&= \bigl((\overline w\star w)\times_{\mathbf{PF}}(\overline u\star u)\bigr)(a_3,b_3,a_1',b_1').
\end{align*}
Since we have checked on every element, we can conclude that; $ (\overline w\times_{\mathbf{PF}}\overline u)\star (w\times_{\mathbf{PF}}u) = (\overline w\star w)\times_{\mathbf{PF}}(\overline u\star u). $

Next, we check naturality of the associator, that is, for each $w:\mathscr{P}_1\to\mathscr P_2$, $u:\mathscr{Q}_1\to\mathscr Q_2$, and
$v:\mathscr R_1\to\mathscr R_2$ we examine whether; \[\alpha^\otimes_{\mathscr P_2,\mathscr Q_2,\mathscr R_2} \star\bigl((w\times_{\mathbf{PF}}u)\times_{\mathbf{PF}}v\bigr) = \bigl(w\times_{\mathbf{PF}}(u\times_{\mathbf{PF}}v)\bigr)\star \alpha^\otimes_{\mathscr{P}_1,\mathscr{Q}_1,\mathscr R_1}. \]
We consider the following inclusions $\mathscr P_i\subseteq \mathbf{Set}(A_i,A_i')$, $\mathscr Q_i\subseteq \mathbf{Set}(B_i,B_i')$, $\mathscr R_i\subseteq \mathbf{Set}(C_i,C_i')$, so that the process functions are functions of the following types $w: A_2\times A_1' \to A_2'\times A_1$, $u: B_2\times B_1' \to B_2'\times B_1$, $v: C_2\times C_1' \to C_2'\times C_1$.
The associator \(\alpha^\otimes_{\mathscr P,\mathscr Q,\mathscr R}: (\mathscr P\times \mathscr Q)\times \mathscr R \to \mathscr P\times (\mathscr Q\times\mathscr R)\) is hence a process function which as a function has type \[ \alpha^\otimes_{\mathscr P,\mathscr Q,\mathscr R}: \bigl(A\times (B\times C)\bigr)\times\bigl((A'\times B')\times C'\bigr) \longrightarrow \bigl(A'\times (B'\times C')\bigr)\times\bigl((A\times B)\times C\bigr) \] and is explicitly defined by $\alpha^\otimes_{\mathscr P,\mathscr Q,\mathscr R}\bigl(a,(b,c),\,(a',b'),c'\bigr):=\bigl(a',(b',c'),\,(a,b),c\bigr)$. We now check the required naturality equation on every possible element $ \Bigl(a_2,(b_2,c_2),\,(a_1',b_1'),c_1'\Bigr)\in \bigl(A_2\times(B_2\times C_2)\bigr)\times\bigl((A_1'\times B_1')\times C_1'\bigr)$. First, we compute $ \alpha^\otimes_{\mathscr P_2,\mathscr Q_2,\mathscr R_2} \star\bigl((w\times_{\mathbf{PF}}u)\times_{\mathbf{PF}}v\bigr)$ by finding the unique fixed point \(\tau\in (A_2\times B_2)\times C_2\) satisfying \[ \tau= (\alpha^\otimes_{\mathscr P_2,\mathscr Q_2,\mathscr R_2})^{(A_2\times B_2)\times C_2} \Bigl( a_2,(b_2,c_2),\, \bigl((w\times_{\mathbf{PF}}u)\times_{\mathbf{PF}}v\bigr)^{(A_2'\times B_2')\times C_2'}(\tau (a_1',b_1'),c_1') \Bigr).\]
Since the right-hand side is independent of the primed argument the fixed point equation reduces to $ \tau = \bigl((a_2,b_2),c_2\bigr)$. Evaluating \(\bigl((w\times_{\mathbf{PF}}u)\times_{\mathbf{PF}}v\bigr)\) at $\tau$ gives 
\begin{align*} 
&\bigl((w\times_{\mathbf{PF}}u)\times_{\mathbf{PF}}v\bigr)^{(A_2'\times B_2')\times C_2'} \bigl((a_2,b_2),c_2,\,(a_1',b_1'),c_1'\bigr) \\ 
&\qquad= \bigl((w^{A_2'}(a_2,a_1'),\,u^{B_2'}(b_2,b_1')),\,v^{C_2'}(c_2,c_1')\bigr), \\[3pt]
&\bigl((w\times_{\mathbf{PF}}u)\times_{\mathbf{PF}}v\bigr)^{(A_1\times B_1)\times C_1}
\bigl((a_2,b_2),c_2,\,(a_1',b_1'),c_1'\bigr) \\
&\qquad=
\bigl((w^{A_1}(a_2,a_1'),\,u^{B_1}(b_2,b_1')),\,v^{C_1}(c_2,c_1')\bigr).
\end{align*}
So that post-multiplication by \(\alpha^\otimes_{\mathscr P_2,\mathscr Q_2,\mathscr R_2}\) returns
\begin{align}
\label{eq:LHS-assoc-nat}
&\Bigl( w^{A_2'}(a_2,a_1'),\; \bigl(u^{B_2'}(b_2,b_1'),\,v^{C_2'}(c_2,c_1')\bigr),\; \bigl(w^{A_1}(a_2,a_1'),\,u^{B_1}(b_2,b_1')\bigr),\; v^{C_1}(c_2,c_1') \Bigr).
\end{align}
Now, we evaluate $ \bigl(w\times_{\mathbf{PF}}(u\times_{\mathbf{PF}}v)\bigr)\star \alpha^\otimes_{\mathscr{P}_1,\mathscr{Q}_1,\mathscr R_1}$, on \(\bigl(a_2,(b_2,c_2),\,(a_1',b_1'),c_1'\bigr)\). In order to compute the contraction we first compute the fixed point \(\rho\) satisfying \[ \rho= \bigl(w\times_{\mathbf{PF}}(u\times_{\mathbf{PF}}v)\bigr)^{A_1\times(B_1\times C_1)} \Bigl( a_2,(b_2,c_2),\, (\alpha^\otimes_{\mathscr{P}_1,\mathscr{Q}_1,\mathscr R_1})^{A_1'\times(B_1'\times C_1')}\bigl(\rho,(a_1',b_1'),c_1'\bigr) \Bigr). \]
In fact, the right-hand-side is independent of $\rho$ since $ (\alpha^\otimes_{\mathscr{P}_1,\mathscr{Q}_1,\mathscr R_1})^{A_1'\times(B_1'\times C_1')} \bigl(\rho,(a_1',b_1'),c_1'\bigr) = \bigl(a_1',(b_1',c_1')\bigr)$, and so the fixed point equation reduces to $ \rho= \bigl( w^{A_1}(a_2,a_1'),\; \bigl(u^{B_1}(b_2,b_1'),\,v^{C_1}(c_2,c_1')\bigr) \bigr)$. Using this fixed-point we can now compute the contraction, where the left-output reduces to \[ \bigl(w\times_{\mathbf{PF}}(u\times_{\mathbf{PF}}v)\bigr)^{A_2'\times(B_2'\times C_2')} \bigl(a_2,(b_2,c_2),\,a_1',(b_1',c_1')\bigr) = \Bigl( w^{A_2'}(a_2,a_1'),\; \bigl(u^{B_2'}(b_2,b_1'),\,v^{C_2'}(c_2,c_1')\bigr) \Bigr), \] and the right-output reduces to \[ (\alpha^\otimes_{\mathscr{P}_1,\mathscr{Q}_1,\mathscr R_1})^{(A_1\times B_1)\times C_1} \bigl(\rho,(a_1',b_1'),c_1'\bigr) = \bigl((w^{A_1}(a_2,a_1'),\,u^{B_1}(b_2,b_1')),\,v^{C_1}(c_2,c_1')\bigr). \]
This completes the point-wise check for naturality of the associator. 

We now consider the naturality of the unitors. Recall that we set $I_{\mathbf{PF}} = \mathbf{Set}(\{\bullet\},\{\bullet\})$ as the monoidal unit. We consider $\mathscr P\subseteq\mathbf{Set}(A,A')$ and $\mathscr Q\subseteq\mathbf{Set}(B,B')$, and check that for every $w:\mathscr P\to\mathscr Q$ then $\rho_{\mathscr Q}\star (w\times_{\mathbf{PF}}1_{I_{\mathbf{PF}}}) = w\star \rho_{\mathscr P}$.
Recall, that as a function we have $w: B\times A' \longrightarrow B'\times A$. For convenience we write $U:=\{\bullet\}$, so that $I\subseteq\mathbf{Set}(U,U)$.

Recall that the right unitor $\rho_{\mathscr P}:\mathscr P\times I\to\mathscr P$ as a function has type $\rho_{\mathscr P}: A\times(A'\times U)\to A'\times(A\times U)$ and is defined component-wise by $\rho_{\mathscr P}\bigl(a,(a',\bullet)\bigr):=\bigl(a',(a,\bullet)\bigr)$.
Similarly, recall that $\rho_{\mathscr Q}: B\times(B'\times U)\to B'\times(B\times U)$,  $\rho_{\mathscr Q}\bigl(b,(b',\bullet)\bigr):=\bigl(b',(b,\bullet)\bigr)$.
We now evaluate $\rho_{\mathscr Q}\star (w\times_{\mathbf{PF}}1_I)$ on an arbitrary element $\bigl(b,(a',\bullet)\bigr)\in B\times(A'\times U)$. First, we must compute the unique fixed point, 
\[ \tau = \rho_{\mathscr Q}^{B\times U} \Bigl( b,\, (w\times_{\mathbf{PF}}1_I)^{B'\times U}(\tau,(a',\bullet)) \Bigr), \]
where \(\rho_{\mathscr Q}^{B\times U}\) is independent of its second argument, and more precisely the fixed point equation simply reduces to $\tau=(b,\bullet)$.

We now evaluate the contraction $\star$ using this \(\tau\) and then apply the definition of $\times_{\mathbf{PF}}$ to reach $
(w\times_{\mathbf{PF}}1_I)^{B'\times U}\bigl((b,\bullet),(a',\bullet)\bigr) = \bigl(w^{B'}(b,a'),\bullet\bigr)$ and $ (w\times_{\mathbf{PF}1_I}^{A\times U})\bigl((b,\bullet),(a',\bullet)\bigr)
  = \bigl(w^{A}(b,a'),\bullet \bigr)$, so that
\begin{align*}
\bigl(\rho_{\mathscr Q}\star (w\times_{\mathbf{PF}}1_I)\bigr)\bigl(b,(a',\bullet)\bigr)
&= \Bigl( \rho_{\mathscr Q}^{B'}\bigl(b,(w^{B'}(b,a'),\bullet)\bigr), (w\times_{\mathbf{PF}}1_I)^{A\times U}\bigl((b,\bullet),(a',\bullet)\bigr) \Bigr)\\
&= \Bigl( w^{B'}(b,a'),\; (w^{A}(b,a'),\bullet) \Bigr).
\end{align*}

Now, we compute \(w\star \rho_{\mathscr P}\). First, we compute the fixed-point by imposing $ \sigma = w^{A}\Bigl( b,\, \rho_{\mathscr P}^{A'}\bigl(\sigma,(a',\bullet)\bigr) \Bigr)$, which forces $\sigma = w^{A}(b,a')$. Then, the composite evaluates to
\begin{align*}
\bigl(w\star \rho_{\mathscr P}\bigr)\bigl(b,(a',\bullet)\bigr)
&= \Bigl( w^{B'}\bigl(b,\rho_{\mathscr P}^{A'}(\sigma,(a',\bullet))\bigr), \; \rho_{\mathscr P}^{A\times U}\bigl(\sigma,(a',\bullet)\bigr) \Bigr)\\
&= \Bigl( w^{B'}(b,a'),\; (w^{A}(b,a'),\bullet) \Bigr),
\end{align*}
Consequently, we have checked point-wise, that indeed
\(\rho_{\mathscr Q}\star (w\times_{\mathbf{PF}}1_I)=w\star\rho_{\mathscr P}\).

Regarding coherence, recall that the functor $(f,g)\longmapsto \mathrm{Sh}_{f,g}$ preserves identities, $\star$-composition, and tensor products of bijections, and furthermore that each of $\alpha^{\otimes},\lambda^{\otimes},\rho^{\otimes},\sigma$
is the image under $\mathrm{Sh}$ of the corresponding coherence bijection in $\mathbf{Set}$.
Consequently, all symmetric monoidal coherence axioms in $\mathbf{PF}$ (pentagon, triangle, involutivity
$\sigma\circ\sigma=1$, and the hexagon axioms) are inherited from those in $\mathbf{Set}$.

Finally, the dualiser can be extended to a full and faithful functor, given $w \in \{\mathscr{P} \rightarrow \mathscr{Q}\}$ we can construct $w^{*}$ by the following series of isomorphisms $\{\mathscr{P} \rightarrow \mathscr{Q}\} \cong  \mathscr{Q} \parr \mathscr{P}^{*} \cong  \mathscr{P}^{*} \parr \mathscr{Q} \cong  \mathscr{P}^{*} \parr \mathscr{Q}^{**} \cong \{\mathscr{Q}^{*} \rightarrow \mathscr{P}^{*} \}$. 
Concretely, for each $ w:\mathscr P\to\mathscr Q$ with $ \mathscr P\subseteq \mathbf{Set}(A,A')$, and $ \mathscr Q\subseteq \mathbf{Set}(B,B')$, recall that $w$ has (as a function) the type $w: B\times A' \longrightarrow B'\times A$. It's dual process function is then the function $w^*: A'\times B \to A\times B'$ given by $w^*(a',b):=\bigl(w^{A}(b,a'),\,w^{B'}(b,a')\bigr)$, or equivalently by  $w^*=\texttt{swap}_{A,B'}\circ w\circ \texttt{swap}_{A',B}$.
To check functoriality, we will check that both $(1_{\mathscr P})^*=1_{\mathscr{P}^*}$, and $(u\star w)^* = w^*\star u^*$.

Regarding identities, first recall that $1_{\mathscr P}=\texttt{swap}_{A,A'}:A\times A'\to A'\times A$, so that $(1_{\mathscr P})^* = \texttt{swap}_{A',A}\circ \texttt{swap}_{A,A'}\circ \texttt{swap}_{A',A} = \texttt{swap}_{A',A} = 1_{\mathscr{P}^*}$.

Regarding the composition law,  we consider $\mathscr R\subseteq \mathbf{Set}(C,C')$ and let $u:\mathscr Q\to\mathscr R$ be another process function, with type $u: C\times B' \to C'\times B$ as a function. Now, fixing $(a',c)\in A'\times C$, to compute $(u\star w)(c,a')$ we must use the unique fixed-point $\tau \in B$ satisfying $\tau = u^{B}\bigl(c,\ w^{B'}(\tau,a')\bigr)$. Using this $\tau$ we have that $(u\star w)(c,a')=\Bigl(u^{C'}\bigl(c,\ w^{B'}(\tau,a')\bigr),\ w^{A}(\tau,a')\Bigr)\in C'\times A$.
Finally, by applying $(-)^*$ we find that $(u\star w)^*(a',c)=\Bigl(w^{A}(\tau,a'),\ u^{C'}\bigl(c,\ w^{B'}(\tau,a')\bigr)\Bigr)$. 

We now compute $(w^*\star u^*)(a',c)$. First, recall that $u^*: B'\times C \to B\times C'$, with $u^*(b',c)=\bigl(u^{B}(c,b'),\ u^{C'}(c,b')\bigr)$, and $w^*: A'\times B \to A\times B'$, with $w^*(a',b)=\bigl(w^{A}(b,a'),\ w^{B'}(b,a')\bigr)$.
To compute the contraction $w^*\star u^*$, we must identify the element $\beta\in B'$ uniquely fixed by the equation
\[
\beta = (w^*)^{B'}\bigl(a',\ (u^*)^{B}(\beta,c)\bigr)
       = w^{B'}\bigl(u^{B}(c,\beta),\ a'\bigr).
\]
Using this fixed-point $\beta$, the contraction evaluates to \[(w^*\star u^*)(a',c) = \Bigl( w^{A}\bigl(u^{B}(c,\beta),a'\bigr),\ u^{C'}(c,\beta) \Bigr). \]

To show equality of the two terms, we first establish a one-to-one correspondence between fixed-points. From the defining fixed-point equation for $\tau$, we set $\beta := w^{B'}(\tau,a') \in B'$, then the equation for $\tau$ becomes $\tau=u^{B}(c,\beta)$, and substituting into the definition of $\beta$ yields $\beta = w^{B'}\bigl(u^{B}(c,\beta),a'\bigr)$. Conversely, if $\beta$ satisfies $\beta=w^{B'}(u^{B}(c,\beta),a')$ and we set $\tau:=u^{B}(c,\beta)$, then $\tau=u^{B}(c,w^{B'}(\tau,a'))$. Substituting $\beta=w^{B'}(\tau,a')$ gives $ (w^*\star u^*)(a',c) = \Bigl( w^{A}(\tau,a'),\ u^{C'}\bigl(c,\ w^{B'}(\tau,a')\bigr) \Bigr)$.
Consequently, $(u\star w)^*(a',c)=(w^*\star u^*)(a',c)$ for all $(a',c)$, and so, $(u\star w)^*=w^*\star u^*$.

We now confirm that in fact $\{\mathscr{P} \times \mathscr{Q} \rightarrow \mathscr{R}\} \cong \{\mathscr{P}\rightarrow( \mathscr{Q} \rightarrow \mathscr{R})\}$. Indeed consider that $\{\mathscr{P} \times \mathscr{Q} \rightarrow \mathscr{R}\} \cong \mathscr{R} \parr (\mathscr{P} \times \mathscr{Q})^{*} \cong   \mathscr{R} \parr( \mathscr{Q}^{*}   \parr \mathscr{P}^{*}) \cong (  \mathscr{R} \parr \mathscr{Q}^{*} )  \parr \mathscr{P}^{*} \cong (\{  \mathscr{Q} \rightarrow \mathscr{R} \}) \parr \mathscr{P}^{*} \cong \{\mathscr{P} \rightarrow ( \mathscr{Q} \rightarrow \mathscr{R})\}$. This proves closure and hence star-autonomy of this category.

\end{proof}

\begin{theorem}
    Taking all objects $A,B,A',B', \dots$ to be first order, the following identifications can be made;
    \begin{itemize}
        \item $A \Rightarrow A' \cong \mathbf{Set}(A,A')$,
        \item $(A \Rightarrow A')  \otimes (B \Rightarrow B')  \cong \mathbf{Set}(A , A' ) \times \mathbf{Set}(B,B')$,
        \item $(A \Rightarrow A')  \parr (B \Rightarrow B')  \cong \mathbf{Set}(A \times B , A' \times B' ) $.
        \end{itemize}
        Consequently, 
        \begin{itemize}
        \item $ (A \Rightarrow \Omega) \parr (S \Rightarrow (S \times \realsgeq)) \cong $ POMDPs,
        \item $ (\parr_i (A_i \Rightarrow \Omega_i)) \parr (S \Rightarrow (S \times \realsgeq)) \cong $ $\dec$-POMDPs,
        \item $ (\otimes_i (A_i \Rightarrow \Omega_i)) \parr (S \Rightarrow (S \times \realsgeq)) \cong $ observation-independent $\dec$-POMDPs.
    \end{itemize}
    Furthermore,
    \begin{itemize}
        \item $ (A \Rightarrow \Omega) \Rightarrow (M \Rightarrow M') \cong $ process functions,
        \item $   \otimes_i ((A_i \Rightarrow A_i') \Rightarrow (B_i \Rightarrow B_i')) \cong $ decentralised process functions,
                \item $ ( \otimes_i (A_i \Rightarrow \Omega_i') ) \Rightarrow (M \Rightarrow M') \cong $ multi-input process functions.
    \end{itemize}
\end{theorem}

\begin{proof}
  First note that $\mathbf{Set}(\{\bullet\} , A) \Rightarrow \mathbf{Set}(\{\bullet\} , A') \cong  \{ \text{process functions } w:  \mathbf{Set}(\{\bullet\} , A) \rightarrow \mathbf{Set}(\{\bullet\}, A') \}$, which is the set of functions $w:\{\bullet\} \times A \rightarrow A' \times \{\bullet\}$ such that for each $f: \{\bullet\} \rightarrow A$ there is a unique solution to $f(w(\bullet, a)) = a$. Every $w$ satisfies this requirement since each such $f$ is a constant, consequently $\mathbf{Set}(\{\bullet\} , A) \Rightarrow \mathbf{Set}(\{\bullet\} , A') \cong \{ \text{process functions } w:  \mathbf{Set}(\{\bullet\} , A) \rightarrow \mathbf{Set}(\{\bullet\}, A') \} \cong  \{ \text{functions } w:  \{\bullet\} \times A \rightarrow A'  \times \{\bullet\} \} \cong  \{ \text{functions } w:  A \rightarrow A' \} \cong  \mathbf{Set}(A,A')$.

  Now note that  $(A \Rightarrow A')  \otimes (B \Rightarrow B') := (A \Rightarrow A')  \times (B \Rightarrow B')   \cong \mathbf{Set}(A , A' ) \times \mathbf{Set}(B,B')$.
  Next we have that $(A \Rightarrow A')  \parr (B \Rightarrow B')  \cong \mathbf{Set}(A , A' ) \parr \mathbf{Set}(B,B')  \cong (\mathbf{Set}(A , A' )^{*} \times \mathbf{Set}(B,B')^{*})^{*}  $ note that $\mathbf{Set}(A , A' )^{*} \cong  \{ \textrm{constant } f: A' \rightarrow A  \}$ and that for every $w \in \mathbf{Set}(A \times B , A' \times B' ) $ and every pair of constants $f,g$ then the equation $(f \times g)(w(a,b))  = (a,b)$ has a unique fixed point. Consequently we have that $(\mathbf{Set}(A , A' )^{*} \times \mathbf{Set}(B,B')^{*})^{*} \cong \mathbf{Set}(A \times B , A' \times B' ) $. 
  
From the characterisation of the $\parr$ is follows immediately that $ (A \Rightarrow \Omega) \parr (S \Rightarrow (S \times \realsgeq)) \cong $ POMDPs, and  $ (\parr_i (A_i \Rightarrow \Omega_i)) \parr (S \Rightarrow (S \times \realsgeq)) \cong $ $\dec$-POMDPs. What remains, to check is that $ (\otimes_i (A_i \Rightarrow \Omega_i)) \parr (S \Rightarrow (S \times \realsgeq)) \cong $ observation-independent $\dec$-POMDPs. First, note that $ (\otimes_i (A_i \Rightarrow \Omega_i)) \parr (S \Rightarrow (S \times \realsgeq))  \cong   (S \Rightarrow (S \times \realsgeq) )^* \Rightarrow  (\otimes_i (A_i \Rightarrow \Omega_i))  $, where $ (S \Rightarrow (S \times \realsgeq) )^* \cong \mathbf{Set}( S , S \times \realsgeq)^*$ which is in turn simply the set of constant functions of type $ (S \times \realsgeq ) \rightarrow S$. Using this, we see that the requirement for membership $\mathcal{P} \in (\otimes_i (A_i \Rightarrow \Omega_i)) \parr (S \Rightarrow (S \times \realsgeq))  $ is that for each element $\mathcal{S} \in S$ and constant function $f_s (s',r) = s$ then the contraction $p_{f_s}(\vec a) = \mathcal{P}^\mathcal{O}(\vec a , s)$ is an element of $ (\otimes_i (A_i \Rightarrow \Omega_i))  :=  (\times_i (A_i \Rightarrow \Omega_i)) $. In other-words, the requirement for membership is precisely the party-wise factorisation of observation-independence.

  Moving on to process functions we have that $ (A \Rightarrow A') \Rightarrow (B \Rightarrow B') \cong  \mathbf{Set}(A , A' )  \Rightarrow \mathbf{Set}(B , B' ) $. By the product factorization for the tensor we have that $ \otimes_i (A_i \Rightarrow A_i') \Rightarrow (B_i \Rightarrow B_i') = \times_i (A_i \Rightarrow A_i') \Rightarrow (B_i \Rightarrow B_i') \cong \times_i  \mathbf{Set}(A_i , A_i' ) \Rightarrow  \mathbf{Set}(B_i , B_i' )$. Finally note that $w \in  \otimes_i  \mathbf{Set}(A_i , A_i' ) \Rightarrow \mathbf{Set}(B , B' )  $ if and only $w$ satisfies the unique fixed point criterion for any product $\times_i f_i \in \times_i \mathbf{Set}(A_i , A_i' )$.
\end{proof}

\section{Truth table for the GYNI game}

The win conditions can be concretely represented in terms of a truth table:
\[\begin{tabular}{ccccccc}
\toprule
$X_1$ & $X_2$ & $X_3$ & $\mathrm{maj}(X_1,X_2,X_3)$ & $Y_1$ & $Y_2$ & $Y_3$ \\
\midrule
0 & 0 & 0 & 0 & 0 & 0 & 0 \\
0 & 0 & 1 & 0 & 1 & 0 & 0 \\
0 & 1 & 0 & 0 & 0 & 0 & 1 \\
0 & 1 & 1 & 1 & 0 & 0 & 1 \\
1 & 0 & 0 & 0 & 0 & 1 & 0 \\
1 & 0 & 1 & 1 & 1 & 0 & 0 \\
1 & 1 & 0 & 1 & 0 & 1 & 0 \\
1 & 1 & 1 & 1 & 0 & 0 & 0 \\
\bottomrule
\end{tabular}.\]

\end{document}